\documentclass[11pt,a4paper]{article}
\pdfoutput=1
\usepackage[hyperref]{acl2021}
\usepackage{times}
\usepackage{latexsym}

\usepackage{microtype}

\usepackage{color}

\newcommand{\orange}[1]{\textcolor{black}{#1}}

\usepackage[mathscr]{eucal}
\newcommand{\Prem}{\mathcal{P}}

\usepackage{CJKutf8}

\usepackage{xcolor,colortbl}
\definecolor{Gray}{gray}{0.9}
\DeclareRobustCommand{\hlgray}[1]{{\sethlcolor{Gray}\hl{#1}}}
\newcommand{\mygray}[1]{\cellcolor{Gray}{#1}}

\usepackage{subcaption}
\usepackage{graphicx}
\usepackage{multirow}
\usepackage{booktabs}
\usepackage{supertabular}
\usepackage{soul}
\usepackage{enumitem}

\usepackage{comment}
\usepackage{graphics}
\usepackage{arydshln}
\usepackage{rotating}
\usepackage{float}
\usepackage{adjustbox}

\usepackage{amssymb}
\usepackage{longtable}
\usepackage{amsfonts,amssymb}
\usepackage{MnSymbol}

\aclfinalcopy %
\def\aclpaperid{3035} %

\title{Investigating Transfer Learning in Multilingual Pre-trained Language Models 
through Chinese Natural Language Inference}

\author{
Hai Hu$^{\dagger\divideontimes}$ \quad He Zhou$^\dagger$ \quad Zuoyu Tian$^\dagger$ \quad Yiwen Zhang$^\dagger$ \quad Yina Ma$^\ddagger$ \\ 
\textbf{Yanting Li}$^\triangleleft$ \quad
\textbf{Yixin Nie}$^\circ$ \quad \textbf{Kyle Richardson}$^\diamond$\\
$^\dagger$Indiana University Bloomington\quad
$^\ddagger$Brigham Young University \quad $^{\divideontimes}$Shanghai Jiaotong Univ.\\
$^\triangleleft$Northwestern University \quad $^\circ$UNC Chapel Hill \quad $^\diamond$Allen Institute for AI 
\\
{
\footnotesize {\tt hu.hai@outlook.com; \{hzh1,zuoytian,yiwezhan,yinama\}@iu.edu; 
}}\\
{
\footnotesize {\tt
yanting.li@northwestern.edu;
yixin1@cs.unc.edu;
kyler@allenai.org
}}
}

\date{}

\begin{document}
\maketitle
\begin{abstract}

Multilingual transformers (XLM, mT5) have been shown to have remarkable transfer skills in zero-shot settings. Most transfer studies, however, rely on automatically translated resources (XNLI, XQuAD), making it hard to discern the particular linguistic knowledge that is being transferred, and the role of expert annotated monolingual datasets when developing task-specific models. We investigate the cross-lingual transfer abilities of XLM-R for Chinese and English natural language inference (NLI), with a focus on the recent large-scale Chinese dataset OCNLI. To better understand linguistic transfer, we created 4 categories of challenge and adversarial tasks (totaling 17 new datasets\footnote{All new datasets/code are released at \urlstyle{rm}\url{https://github.com/huhailinguist/ChineseNLIProbing}. }) for Chinese that build on several well-known resources for English (e.g., \emph{HANS}, NLI \emph{stress-tests}). We find that cross-lingual models trained on English NLI do transfer well across our Chinese tasks (e.g., in 3/4 of our challenge categories, they perform as well/better than the best monolingual models, even on 3/5 uniquely Chinese linguistic phenomena such as \emph{idioms, pro drop}). These results, however, come with important caveats: cross-lingual models often perform best when trained on a mixture of English and high-quality monolingual NLI data (OCNLI), and are often hindered by automatically translated resources (XNLI-zh). For many phenomena, all models continue to struggle,  highlighting the need for our new diagnostics to help benchmark Chinese and cross-lingual models.

\end{abstract}

\section{Introduction}

Recent pre-trained multilingual transformer models, such as XLM(-R) \cite{conneau2019cross,xlm-r}, mT5 \cite{mt5} and others \cite{liu2020multilingual,lewis2020pre} have been shown to be  successful in NLP tasks for several non-English languages \citep{parsinlu,choi2021analyzing},
as well as in multilingual benchmarks \citep{bert,xlm-r,mt5,artetxe-etal-2020-cross}. A particular appeal is that they can be used for \emph{cross-lingual} and \emph{zero-shot transfer}. That is, after pre-training on a raw, unaligned corpus consisting of text from many languages,  models can be subsequently fine-tuned on a particular task in a resource-rich language (e.g., English) and directly applied to the same task in other languages without requiring any additional language-specific training. 

Given this recent progress, a natural question arises: does it make sense to invest in large-scale task-specific dataset construction for low-resourced languages, or does cross-lingual transfer alone suffice for many languages and tasks? A closely related question is: how well do multilingual models transfer across specific linguistic and language-specific phenomena? While there has been much recent work on probing multilingual models \cite{wu2019mbert,pires2019mbert,karthikeyan2019mbert}, \emph{inter alia}, a particular limitation is that most studies rely on automatically translated resources such as XNLI \cite{xnli} and XQuAD \cite{artetxe-etal-2020-cross}, which makes it difficult to discern the particular linguistic knowledge that  is  being  transferred  and  the  role of large-scale, expert annotated monolingual datasets when building task- and language-specific models.

In this paper, we investigate the cross-lingual transfer abilities of XLM-R \cite{xlm-r} for Chinese natural language inference (NLI). Our focus on Chinese NLI is motivated by the recent release of the first large-scale, human-annotated Chinese NLI dataset OCNLI (\textbf{O}\emph{riginal} \textbf{C}\emph{hinese} \textbf{NLI}) \cite{ocnli}\footnote{To our knowledge, OCNLI is currently the largest non-English NLI dataset that was annotated in the style of English MNLI without any translation.}, which we use to directly investigate the role of high-quality task-specific data vs. English-based cross-lingual transfer. To better understand linguistic transfer, and help benchmark recent SOTA Chinese NLI models, we created 4 categories of challenge/adversarial tasks (totaling 17 new datasets) for Chinese that build on  several  well-established resources for English and the literature on model probing (see \citet{poliak2020survey}). Our new resources, which are summarized in Table~\ref{tab:sum:4sets}, include: a new set of diagnostic tests in the style of the SuperGLUE ~\citep{superglue} and CLUE \cite{clue} diagnostics; Chinese versions of the HANS dataset \cite{hans} and NLI stress-tests \cite{naik2018stress}, as well as a collection of the basic reasoning and logic \emph{semantic probes} for Chinese based on \citet{probing2020}. 

\begin{table}[t]
\centering
\footnotesize
\scalebox{.85}{
\begin{tabular}{lll}\toprule
 & category & n  \\\midrule
\multirow{2}{*}{\rotatebox[origin=c]{90}{\parbox{.8cm}{Chinese\\HANS}}} & \parbox{2cm}{Lexical overlap} & 1,428   \\
 & Subsequence & 513   \\\midrule
\multirow{5}{*}{\rotatebox[origin=c]{90}{stress tests}} 
& Distraction: 2 categories & 8,000   \\
 & Antonym & 3,000   \\
 & Synonym & 2,000   \\
 & Spelling & 11,676  \\
 & Numerical reasoning & 8,613   \\\midrule
\multirow{7}{*}{\rotatebox[origin=c]{90}{diagnostics}} & CLUE~\cite{clue} & 514   \\
 & CLUE expansion (ours) & 796   \\
 & World knowledge (ours) & 38   \\
 & Classifier (ours) & 139   \\
 & Chengyu/idioms (ours) & 251   \\
 & Pro-drop (ours) & 198   \\
 & Non-core arguments (ours) & 186 \\\midrule
\multirow{6}{*}{\rotatebox[origin=c]{90}{\parbox{1.5cm}{semantic\\probing}}} & Negation & 1,002   \\
 & Boolean & 1,002   \\
 & Quantifier & 1,002   \\
 & Counting & 1,002   \\
 & Conditional & 1,002   \\
 & Comparative & 1,002\\\midrule
sum & & 43,364\\\bottomrule  
\end{tabular}
}
\caption{Summary statistics of the four evaluation sets. \label{tab:sum:4sets}}
\end{table}

Our results are largely positive: We find that cross-lingual models trained exclusively on English NLI do transfer relatively well across our new Chinese tasks (e.g., in 3/4 of the challenge categories shown in Table~\ref{tab:sum:4sets}, they perform overall as well or better than the best monolingual Chinese models without additional specialized training on Chinese data, and have competitive performance on OCNLI).  A particularly striking result is that such models even perform well on 3/5 uniquely Chinese linguistic phenomena such as \emph{idioms, pro drop}, providing evidence that many language-specific  phenomena do indeed transfer. These results, however, come with important caveats: on several phenomena we find that models continue to struggle and are far outpaced by conservative estimates of human performance (e.g., our best model on Chinese HANS remains $\sim$19\% behind human performance), highlighting the need for more language-specific diagnostics tests. Also, fine-tuning models on mixtures of English NLI data and high-quality monolingual data (OCNLI) consistently performs the best, whereas mixing with automatically translated datasets (XNLI-zh) can greatly hinder model performance. This last result shows that high-quality monolingual datasets still play an important role when building cross-lingual models, however, the particular type of monolingual dataset that is needed can vary and is best informed by targeted behavioral testing of the type we pursue here.

\section{Related Work}

There has been a lot of work on trying to understand  multilingual transformers \cite{wu2019mbert,pires2019mbert}, which has focused on either examining the representation of different layers in the transformer architecture or the lexical overlap between languages. \citet{karthikeyan2019mbert} investigate the role of network depth and number of attention heads, as well as
syntactic/word-order similarity on the cross-lingual transfer performance. 
In addition to studies cited at the outset, positive results of cross-lingual transfer across a wide range of languages are reported in \citet{wu2020all,nozza2020mask}, with a focus on transfer across specific tasks such as POS tagging, NER; in contrast, we focus on different categories of linguistic transfer, which has received less attention, as well as the role of monolingual data for transfer in NLI.

Studies into the linguistic abilities and robustness of current NLI models have proliferated in recent years, partly owing to the discovery of systematic biases, or \emph{annotation artifacts} \cite{gururangan2018annotation,poliak2018hypothesis}, in benchmark NLI datasets such as SNLI \cite{snli} and MNLI \cite{mnli}. This has been coupled with the development of new \emph{adversarial tests} such as  \emph{HANS} \cite{hans} and the NLI \emph{stress-tests} \cite{naik2018stress}, as well as several new linguistic \emph{challenge datasets} \cite{breaknli,probing2020,geiger-etal-2020-neural,yanaka2019help,saha2020conjnli,goodwin2020ystematicity}, \emph{inter alia}, that focus on a wide range of linguistic and reasoning phenomena. All of this work focuses exclusively on English, whereas we focus on constructing analogous probing datasets tailored to Chinese to help advance research on Chinese NLI and cross-lingual transfer.

There has been a surge in the development
of NLI resources for languages other than English. Such resources are often created in the following two ways:
(1) from \emph{scratch}, in the style of MNLI \cite{mnli}, where annotators are used to produce hypotheses and inference labels based on a provided set of premises, as pursued for Chinese OCNLI \cite{ocnli}, or SciTail \cite{scitail}, where sentences are paired automatically and labeled by annotators \cite{amirkhani2020farstail,hayashibe2020japanese}. (2) Through \emph{automatic} \cite{xnli,turkish-nli,assin2} or \emph{manual} \cite{sick-dutch} translation from existing English datasets. Studies on cross-lingual transfer for NLI have largely focused on XNLI \cite{xnli}, which we show has limited utility for Chinese NLI transfer.

\section{Dataset creation}

In this section, we describe the details of the 4 types of challenge datasets we constructed for Chinese to study cross-lingual transfer (see details in Table~\ref{tab:sum:4sets}). They fit into two general categories: \textbf{Adversarial datasets} (Section~\ref{sec:ad}) built largely from patterns in OCNLI~\cite{ocnli} and XNLI~\cite{xnli} and \textbf{Probing/diagnostic datasets} (Section~\ref{sec:prob}), which are \orange{built from scratch in a parallel fashion to existing datasets in English}. 

While we aim to mimic the annotation protocols pursued in the original English studies,  we place the additional methodological constraint that each new dataset is vetted, either through human annotation using a disjoint set of Chinese linguists, or through internal mediation among local Chinese experts; details are provided below.

\subsection{Adversarial dataset}
\label{sec:ad}

Examples from the 7 adversarial tests we created are illustrated in Table~\ref{tab:ex:hans:stress}.\footnote{A more detailed description of
the data creation process
can be found in Appendix~\ref{sec:appendix:ex}.} Chinese HANS is built from patterns extracted in the large-scale Chinese NLI dataset OCNLI \cite{ocnli}, whereas the \textbf{Distraction}, \textbf{Antonym}, \textbf{Synonym} and \textbf{Spelling} subsets are built from an equal mixture of OCNLI and XNLI-zh \cite{xnli} data; in the latter case, such a difference allows us to fairly compare the effect of training on expert-annotated (i.e., OCNLI) vs. automatically translated data (i.e., XNLI-zh) as detailed in Section~\ref{sec:exp}.

\paragraph{Chinese HANS}

\citet{hans} discovered systematic biases/heuristics in the MNLI dataset, 
which they named ``lexical/subsequence/constituent'' overlap.
``Lexical overlap'' is defined to be the pairs where the vocabulary of the hypothesis
is a subset of the vocabulary of the premise.
For example, ``\textit{The boss is meeting the client.}'' and ``\textit{The client is meeting the boss.}'',
which has an entailment relation.
However, lexical overlap does not necessarily mean the premise
will entail the hypothesis, e.g., ``\textit{The judge was paid by the actor.}'' does not entail ``\textit{The actor was paid by the judge.}'' (examples from \citet{hans}). Thus a model relying on the heuristic will fail catastrophically in the second case.

Inspired by the English HANS, we examine whether OCNLI also possesses
such biases, as it has a similar annotation procedure as MNLI. 
We follow the design of the original HANS experiments, and adapt
their scripts\footnote{\url{https://github.com/tommccoy1/hans}} to extract examples in OCNLI that satisfy the two heuristics. We find a heavy bias towards ``entailment'', where 79.5\% of such
examples are ``entailment'', similar to MNLI. 
To construct a Chinese HANS, we first look into syntactic structures of the examples having the two heuristics. Then we write 29 templates for the \textit{lexical overlap} heuristic and 11 templates for  \textit{subsequence overlap}.\footnote{For details of the templates, see Appendix~\ref{sec:app:hans:templates}.
}
Using the templates and a vocabulary 
of 263 words, we generated
1,941 NLI pairs. See Table~\ref{tab:ex:hans:stress} for examples and Appendix~\ref{sec:appendix:ex} for details.

\begin{CJK*}{UTF8}{gbsn}
\begin{table*}[t]
\scriptsize
\centering 
\scalebox{.95}{
\begin{tabular}{p{.2cm}p{1cm}p{.3cm}p{7cm}p{4.5cm}p{.3cm}}\toprule
 & category & n & premise & hypothesis & label \\\midrule
\multirow{2}{*}{\rotatebox[origin=c]{90}{\parbox{.8cm}{Chinese \\ HANS}}} & Lexical overlap & 1428 & 我们把银行职员留在电影院了。  We left the bank clerk in the cinema. &银行职员把我们留在电影院了。  The bank clerk left us in the cinema. & C \\
 & Subsequence & 513 & 谁说\underline{律师都是穿西装的}。  Who told you that \ul{all lawyers wear suits}. & \underline{律师都是穿西装的}。   \ul{All lawyers wear suits}. & C \\\midrule
\multirow{6}{*}{\rotatebox[origin=c]{90}{\parbox{3cm}{stress tests}}} & Distraction \newline (add to\newline premise) & 4000 & 国有企业改革的思路和方针政策已经明确,而且刚\underline{做完手术出院的病人不应剧烈运动}。 The policy of the reform of state-owned enterprises is now clear, \ul{and patients who just had surgery shouldn't have intense exercise}. & 根本不存在国有企业。   The state-owned enterprises don't exist. & C \\
 & Distraction \newline (add to\newline hypothesis) & 4000 & 这时李家院子挤满了参观的人。  During this time, the Li family's backyard is full of people who came to visit. & 这地方有个姓李的人家, \underline{而且真的不是假的}。 There is a Li family here, and \ul{true is not false}. & E \\
 & Antonym & 3000 & 一些地方财政收支矛盾较大。    The disagreement about local revenue is relatively big. & 一些地方财政收支矛盾较小。  The disagreement about local revenue is relatively small. & C \\
 & Synonym & 2000 & 海部组阁困难说明了什么。  What can you tell from the \underline{difficulties} from Kaifu's attempt to set up a cabinet? & 海部组阁艰难说明了什么。  What can you tell from the \underline{hardships} from Kaifu's attempt to set up a cabinet? & E \\
 & Spelling & 2980 & 身上裹一件工厂发的棉大衣,手插在袖筒里。 (Someone is) wrapped up in a big cotton coat the factory gave with hands in the sleeves & 身上\underline{质}少一件衣服。 There's at \underline{least} [typo] one coat on the body. & E \\
 & Numerical\newline reasoning & 8613 & 小红每分钟打不到510个字。 Xiaohong types fewer than 510 words per min. & 小红每分钟打110个字。 Xiaohong types 110 words per min. & N \\\bottomrule
 \end{tabular}}
\caption{Example NLI pairs in Chinese HANS and stress tests with translations. \label{tab:ex:hans:stress}}
\end{table*}

\paragraph{Distraction}
We add distractions to the premise 
or hypothesis, similar to the ``length
mismatch'' and ``word overlap'' conditions
in the NLI \emph{stress tests} of  \citet{naik2018stress}. The distractions are either tautologies (``true is not false'') or a true statement from our world knowledge (``Finnland is not a permanent  member of the UN security council''), which should
not influence the inference label. 
We control whether the distraction 
contain a negation or not, and thus create
four conditions: \textit{premise-negation}, \textit{premise-no-negation}, \textit{hypothesis-negation}, and \textit{hypothesis-no-negation}. See Table~\ref{tab:ex:hans:stress} for examples.

\paragraph{Antonym} We replace a word in the premise with its antonym to form a contradiction. To ensure the quality
of the resulting NLI pairs, 
we manually examine the initially generated data and decided to only 
replace nouns and adjectives, as 
they are more likely to produce
real contradictions. 

\paragraph{Synonym} We replace a word in the premise with its synonym to form an
entailment.

\paragraph{Spelling} We replace one random character in the hypotheses with its homonym (character with the same \textit{pinyin} pronunciation ignoring tones) as this is one of the most common types of misspellings in Chinese. 

\paragraph{Numerical reasoning} We create a probing set for numerical reasoning,
following simple heuristics such as the following. 
When the premise is \textit{Mary types x words per minute}, the entailed hypothesis can be: \textit{Mary types less than y words per minute}, where \textit{x} $<$ \textit{y}.
A contradictory hypothesis:
\textit{Mary types y words per minute}, where \textit{x} $>$ \textit{y} or \textit{x} $<$ \textit{y}.
Then a neutral pair can be produced by reversing the premise and hypothesis of the above entailment pair.
4 heuristic rules (with 6 words for quantification) are used and the seed sentences are 
extracted from Ape210k \citep{ape210k}, a dataset of Chinese elementary-school math problems.
The resulting data contains 8,613 NLI pairs.

For \textbf{quality control} and to compute human performance, we randomly
sampled 50 examples from all subsets
and asked 5  Chinese speakers to verify. Our goal is to mimic the human annotation protocol from \citet{glue-human}, which gives us a \emph{conservative} estimate of human performance given that our annotators received very little instructions. Their majority vote agrees with the gold label
90.0\% of the time, which suggests that our data is of high quality and allows us to later compare against model performance.\footnote{Specifically: 98.0\% on Chinese HANS, 86.0\% on the stress tests. For comparison, different subsets of the English stress tests receives 85\% to 98\% agreement~\cite{naik2018stress}.  }

\subsection{Probing/diagnostic datasets}
\label{sec:prob}
While the Chinese HANS and stress tests
are designed to adversarially test
the models, we also create probing or diagnostic datasets
which are aimed at examining 
the models' linguistic and reasoning abilities. 

\paragraph{Hand-crafted diagnostics} \label{sec:new:diagnostics}

We expanded the diagnostic dataset from the Chinese NLU Benchmark (CLUE) \citep{clue} in the following two ways: 

\begin{CJK*}{UTF8}{gbsn}
First, 6 Chinese linguists (PhD students) created diagnostics for 4 Chinese-specific linguistic phenomena. Here are two of the phenomena:\footnote{For the other two, please refer to Appendix~\ref{sec:appendix:ex}.}
(1) \textit{pro-drop}: subjects or objects in Chinese can be dropped when they can be recovered from the context \citep{li1981mandarin}. Thus the model needs to figure out the subject/object from the context.
(2) \textit{four-character idioms} (i.e., 成语~\textit{Chengyu}). They are a special type of Chinese idioms that has exactly four characters, usually with a figurative meaning different from the literary meaning, e.g., 打草惊蛇~\textit{hit hay startle snake} (behaving carelessly and causing your enemy to become vigilant). \orange{We construct examples to test whether models understand the figurative meaning in the idioms. Specifically, we first create a premise $\Prem$ which includes the idiom, where there is enough context so that a human is highly likely to interpret the idiom figuratively. Then we create an entailed hypothesis that is based on the figurative (correct) interpretation, and a neutral/contradictory hypothesis that uses the literal (incorrect) meaning (see Table~\ref{app:tab:ex:diagnostics} in the Appendix for an example). For each $\Prem$ we write 3 hypothesis, one for each inference relation.  }
We also added diagnostics involving world knowledge.
\end{CJK*}

Second, we double the number of diagnostic pairs for all 9 existing linguistic phenomena in CLUE with pairs whose 
premises are selected from a large news corpus\footnote{We use the BCC corpus \cite{bcc}: \urlstyle{rm} \url{http://bcc.blcu.edu.cn/}.} and hypotheses are hand-written by our linguists, to accompany the 514 artificially created data in CLUE. 
The resulting new diagnostics is 4 times as large as the original one,
with a total of 2,122 NLI pairs. 
For quality control, each pair is
double-checked by local Chinese linguists not involved in this study and the controversial cases
were discarded after a discussion among the 6 linguists.
See Table~\ref{app:tab:ex:diagnostics} in Appendix~\ref{sec:appendix:ex} for examples.

\paragraph{Semantic fragments} 

Following \citet{probing2020} and \citet{salvatore2019logical}, we design synthesized fragments to examine models' understanding
ability of six types of linguistic and logic
inference: \textbf{boolean}, \textbf{comparative}, \textbf{conditional}, \textbf{counting}, \textbf{negation} and \textbf{quantifier}, where
each category has 2-4 templates. See example templates and NLI pairs in Table~\ref{tab:ex:probing}.

The data is generated using context-free grammar rules 
and a vocabulary of 80,000 person names (Chinese and transliterated), 8659 city names and expanded predicates and comparative relations in \citet{probing2020} to
make the data more challenging.
As a result, we generated 
1,000 examples for each fragment. 
For quality control, each template
was checked by 3 linguists/logicians; also 20 examples 
from each category were checked for correctness by local experts.

\begin{CJK*}{UTF8}{gbsn}
\begin{table*}[t]
\scriptsize
\centering 
\scalebox{.95}{
\begin{tabular}{p{1.3cm}p{8cm}p{5cm}l}\toprule
category & premise & hypothesis & label \\\midrule
Negation & 库尔图尔只到过湛江市麻章区，丰隆格只到过大连市普兰店区……
 \textit{person$_1$} only went to \textit{location$_1$}; \textit{person$_2$} only went to \textit{location$_2$}; .... & 库尔图尔没到过大连市普兰店区。\newline \textit{person$_1$} has not been to \textit{location$_2$}. & E \\
Boolean & 何峥、管得宽、李国柱……只到过临汾市襄汾县。 \newline\textit{person$_1$}, \textit{person$_2$} ... have only been to \textit{location$_1$}. & 何峥没到过遵义市红花岗区。 \newline \textit{person$_1$} has not been to \textit{location$_2$}. & E \\
Quantifier  & 
有人到过每一个地方，拥抱过每一个人。\newline Someone has been to every place and hugged every person. & 
王艳没拥抱过包一。  \textit{person$_1$} hasn't hugged \textit{person$_2$}. & N \\
Counting  & 韩声雄只拥抱过罗冬平、段秀芹……赵常。 \newline \textit{person$_1$} only hugged \textit{person$_2$}, \textit{person$_3$} ... \textit{person$_8$}.
 &  韩声雄拥抱过超过10个人。\newline \textit{person$_1$} hugged more than 10 people. & C \\
Conditional & 
……，穆肖贝夸到过赣州市定南县， 如果穆肖贝夸没到过赣州市定南县， 那么张本伟到过呼伦贝尔市阿荣旗。... \textit{person$_n$} has been to \textit{location$_n$}. If \textit{person$_n$} hasn't been to \textit{location$_n$}, then \textit{person$_m$} has been to \textit{location$_m$}. & 
张本伟没到过呼伦贝尔市阿荣旗。 \textit{person$_m$} hasn't been to \textit{location$_m$}. & N \\
Comparative  & 龙银凤比武书瑾、卢耀辉……奈德哈特都小，龙银凤和亚厄纳尔普一样大。\textit{person$_1$} is younger than  \textit{person$_2$},  ...,  \textit{person$_n$};  \textit{person$_1$} is as old as  \textit{person$_m$}  & 亚厄纳尔普比梁培娟大。  \textit{person$_m$} is older than  \textit{person$_{n-2}$}. & C \\\bottomrule
\end{tabular}}
\caption{Example NLI pairs for semantic/logic probing with translations. Each label for each category has 2 to 4 templates; we are only showing 1 template for 1 label. 1,000 examples are generated for each category. \label{tab:ex:probing}}
\end{table*}
\end{CJK*}

\section{Experimental setup}
\label{sec:exp}

Our main goal is to test
whether cross-lingual transfer are
robust against the adversarial and
probing data we created \orange{when evaluated without additional training}. 
Thus we need to compare the best Chinese
monolingual models
with the best multilingual models \orange{trained either on English NLI data alone, or on combinations of Chinese and English data.}\footnote{We also run the same experiments
for Chinese-to-English transfer, 
i.e., fine-tuning XLM-R with OCNLI and evaluate
on the four English counterpart datasets.
We find that transferring 
from OCNLI to English does
not perform as well as monolingual 
English models, likely due to the small
size of OCNLI. Detailed results are reported in Appendix~\ref{sec:app:ch2en:transfer}.}

\paragraph{Chinese monolingual models}
We experimented with two current state-of-the-art transformer models:
RoBERTa-large \citep{roberta} and Electra-large-discriminator \citep{electra}.
We use the Chinese models released from
\citep{cui2020revisiting}\footnote{We use \texttt{hfl/chinese-roberta-wwm-ext-large} from \urlstyle{rm}\url{https://github.com/ymcui/Chinese-BERT-wwm}
and \texttt{hfl/chinese-electra-large-discriminator} from  \urlstyle{rm}\url{https://github.com/ymcui/Chinese-ELECTRA}. }  
implemented the Huggingface Transformer library \cite{transformers}. 

\paragraph{Multilingual model} We use XLM-RoBERTa-large \citep{xlm-r}. 
We choose XLM-R over mT5~\citep{mt5} because XLM-R generally performs better than mT5 under the same model size (see original paper for details). Also, XLM-R as a RoBERTa model is most related architecturally to existing Chinese pre-trained models.

\paragraph{Fine-tuning data for Chinese models \& XLM-R}
      (1) \underline{XNLI}: the full Chinese training set in the machine-translated XNLI dataset, with 390k examples \citep{xnli}. 
      (2) \underline{XNLI-small}: 50k examples from XNLI, the same size as the training data of OCNLI. 
      (3) \underline{OCNLI}: Original Chinese NLI dataset \citep{ocnli}. It is a Chinese NLI dataset collected
      from scratch, following the MNLI procedure, with 50k training examples. We use this to measure the effect of
      the quality of training data; that is, whether it is better to use
      small, high-quality training data (OCNLI), or large, low-quality MT data (XNLI). 
      (4) \underline{OCNLI + XNLI}: a combination of the two training sets, 440k examples. 

\paragraph{Fine-tuning data for XLM-R}
To examine cross-lingual transfer, we finetune XLM-R on English NLI data alone and
English + Chinese NLI data: 
      (1) \underline{MNLI}: 390k examples from MNLI.train \citep{mnli}. 
      (2) \underline{English all NLI}: we combine MNLI \citep{mnli}, SNLI \citep{snli}, FeverNLI \citep{fever,nie2019combining} with ANLI \citep{anli}, a total of 1,313k examples.
      (3) \underline{OCNLI + English all NLI}. 
      (4) \underline{XNLI + English all NLI}. These two are set to examine whether combining Chinese and English
      fine-tuning data is helpful.

We fine-tune the models on OCNLI-dev.
Acknowledging that different training runs can produce very different checkpoints for behavioral testing \cite{d2020underspecification},  we run 5 models on different seeds and report the mean accuracy of the models with the best hyper-parameter setting (for details see Appendix~\ref{sec:app:finetune}).

\section{Results and discussion}

\subsection{Results on OCNLI\_dev}

\begin{table}[t]
\footnotesize
\centering
\scalebox{.85}{
\begin{tabular}{llcc}\toprule
Model & Fine-tuned on & Acc & Scenario \\\midrule
RoBERTa & zh MT: XNLI-small & 67.44 & monolingual \\
RoBERTa & zh MT: XNLI & 70.29 & monolingual \\
RoBERTa & zh ori: OCNLI & 79.11 & monolingual \\
RoBERTa & zh: OCNLI + XNLI & 78.43 & monolingual \\ \hdashline %
XLM-R & zh MT: XNLI & 72.55  & monolingual \\
XLM-R & zh ori: OCNLI & 79.24 & monolingual \\
XLM-R & zh: OCNLI + XNLI & \cellcolor{Gray}{80.31} & monolingual \\ \midrule %
XLM-R & en: MNLI & 71.98 & zero-shot \\
XLM-R & en: En-all-NLI & \cellcolor{Gray}{73.73} & zero-shot \\ \midrule %
XLM-R & mix: OCNLI + En-all-NLI & \cellcolor{Gray}{\textbf{82.18}} & mixed \\
XLM-R & mix: XNLI + En-all-NLI & 74.12  & mixed  \\\bottomrule
\end{tabular}
}
\caption{Results on OCNLI dev.  ``\textbf{Scenario}'' indicates whether the model is fine-tuned on Chinese \emph{only} data (\textbf{monolingual}), English data (\textbf{zero-shot}) or \textbf{mixed} English and Chinese data; results in \hlgray{gray} show best performance for each scenario. Best overall result in \textbf{bold}. Same below.  \label{tab:ocnli:dev}}
\end{table}

Results on the dev set of OCNLI are presented in Table~\ref{tab:ocnli:dev}.
For monolingual RoBERTa, we see a similar performance
as reported in the OCNLI paper \citep{ocnli},
with 79.11\% accuracy. The monolingual Electra
achieves a very close accuracy of 79.02\% (not shown 
in the Table). 
As we see the same trend in the following experiments,
we will therefore only report results on RoBERTa.

For XLM-R, fine-tuning on MNLI or En-all-NLI gives us 
reasonable results of around 72\% to 74\%, which is better than models fine-tuned on XNLI,
indicating that fine-tuning on an English data (MNLI) alone
can outperform monolingual models fine-tuned
on the same data but machine-translated into Chinese (XNLI).\footnote{For these experiments we also tested with another Chinese machine-translated MNLI (CMNLI), translated
by a different MT system, which was released by CLUE \urlstyle{rm} (\url{https://github.com/CLUEbenchmark/CLUE}), and obtained similar results.} 
This is consistent with previous results on 
Korean \cite{choi2021analyzing} and 
Persian \cite{parsinlu} for other NLU tasks. 

What is also interesting is that 
combining OCNLI and En-all-NLI gives us a boost
of 2\% to 82.18\% (a result that is comparable to the current published SOTA),
showing the power of mixing 
high-quality English and Chinese training data.

\subsection{Chinese HANS}

Table \ref{hans-result} shows results of the Chinese HANS data tested on the aforementioned monolingual models and cross-lingual model.

\begin{table*}[t]
\centering
\footnotesize
\scalebox{0.85}{
\begin{tabular}{ll|c|cccc | c }
\toprule
Model & Fine-tuned on & Overall & Lexical Overlap & Sub-sequence & Entailment & Non-Entailment & $\Delta$ \\\midrule
RoBERTa                   & zh MT: XNLI-small                     & 49.48    & 58.12           & 25.42       & 99.22      & 30.26          & 37.18 \\ 
RoBERTa                   & zh MT: XNLI                           & 60.80    & 68.99           & 38.01       & 99.74      & 45.76          & 24.53 \\

RoBERTa                   & zh ori: OCNLI                          & \cellcolor{Gray}{71.72}    & \cellcolor{Gray}{75.39}           & \cellcolor{Gray}{61.48}       & 99.67      & \cellcolor{Gray}{60.91}          & 18.20 \\
RoBERTa                   & zh: OCNLI+XNLI                     & 69.33    & 74.73           & 54.27       & 99.89      & 57.51          & 20.92  \\
\hdashline %
XLM-R                     & zh MT: XNLI                           & 57.74    & 66.47           & 33.45       & 99.96      & 41.42          & 31.13  \\
XLM-R                     & zh ori: OCNLI                          & 61.82    & 65.83           & 50.68       & 99.89      & 47.11          & 32.13  \\
XLM-R                     & zh: OCNLI+XNLI                     & 70.31    & 74.25           & 59.34       & \cellcolor{Gray}{100.00}     & 58.84 & 21.47 \\
\midrule %
XLM-R                     & en: En-all-NLI                     & \cellcolor{Gray}{69.56}    & \cellcolor{Gray}{77.62}           & 47.13       & \cellcolor{Gray}{100.00}     & \cellcolor{Gray}{57.80}          & 15.93 \\
XLM-R                     & en: MNLI                           & 66.74    & 73.12           & \cellcolor{Gray}{48.97}       & 100.00     & 53.89          & 18.09  \\ \midrule %
XLM-R                     & mix: OCNLI+En-all-NLI              & \cellcolor{Gray}{\textbf{78.82}}    & \cellcolor{Gray}{\textbf{81.57}}           & \cellcolor{Gray}{\textbf{71.15}}       & \cellcolor{Gray}{\textbf{100.00}}     & \cellcolor{Gray}{\textbf{70.63}}          & 11.55  \\
XLM-R                     & mix: XNLI+En-all-NLI              & 66.90    & 76.25           & 40.90       & 99.93     & 41.89          & 32.23   \\\midrule
Human & & 98.00& & & &  \\
\bottomrule
\end{tabular}}
\caption{Accuracy on Chinese HANS. $\Delta =$ the difference of accuracy between OCNLI dev and Non-Entailment. %
}
\label{hans-result}
\end{table*}

\textbf{Cross-lingual transfer achieves strong results}.
We first notice that when XLM-R is fine-tuned solely on the English 
data (En-all-NLI), the performance ($\sim$69\%) is only slightly 
worse than the best monolingual model ($\sim$71\%). 
This suggests that cross-lingual transfer from
English to Chinese is quite successful
for an adversarial dataset like HANS.
Second, adding OCNLI to En-all-NLI in the training data
gives a big boost of about 9\%, and achieves
the overall best result. 
This is about 12\% higher than combining XNLI
and the English data, demonstrating
the advantage of the expert-annotated OCNLI over machine translated XNLI, even
though the latter is about 8 times the size 
of the former.
Despite these results, however, we note that all models continue to perform below human performance, suggesting more room for improvement.

Our results also suggest that examples involving the \textit{sub-sequence} heuristics are more difficult than those targeting the \textit{lexical overlap} heuristics for the transformers models we tested (see the ``sub-sequence'' and ``lexical overlap'' columns in Table~\ref{hans-result}). This is in line with the results reported in the English HANS paper (specifically Table 15 in \citet{hans} which also shows that the sub-sequence examples are more difficult for the English BERT model).
Second, for the sub-sequence heuristics, results from monolingual model are 12\% higher than those from XLM-R under the zero-shot transfer setting (61.48\% versus 48.79\% in ``sub-sequence'' column in Table~\ref{hans-result}).
This stands in contrast with the lexical overlap heuristic, where the best monolingual model performs similarly to the best zero-shot cross-lingual transfer (75.39\% versus 77.62\%).
This is one of the few cases where cross-lingual transfer under-performs the monolingual setting by a large margin, suggesting that in certain situations monolingual models may be preferred.

\subsection{Stress tests}

Table~\ref{tab:stress} presents the accuracies
on all the stress tests. We first see that cross-lingual zero-shot transfer using
all English NLI data
performs even better than the best monolingual model ($\sim$74\% vs.~$\sim$71\%).
This demonstrates the power of the cross-lingual transfer-learning. 
Adding OCNLI to all English NLI gives another 
increase of about 3 percentage points (to 77\%), while 
adding XNLI hurts the performance, again showing
the importance of having expert-annotated 
language-specific data.

\begin{table*}[t]
\small
\centering
\scalebox{.82}{
\begin{tabular}{ll|c|cccccccc}
\toprule
Model   & Fine-tuned on             & Overall         & Ant. & Syn. & \parbox{1.2cm}{\centering Distr H} &
\parbox{1.2cm}{\centering Distr H-n} & \parbox{1.2cm}{\centering Distr P} & 
\parbox{1.2cm}{\centering Distr P-n}  & 
\parbox{1cm}{\centering Spell-\\ing} & 
num. 
\\\midrule
RoBERTa & zh MT: XNLI-small       & 59.41 & 43.38   & 99.64   & 51.61 & 51.41 & 70.66  & 71.19            & 69.93    & 28.70      \\
RoBERTa & zh MT: XNLI             & 66.22 & 52.28   & 99.79   &54.83 & 53.8  & 74.55  & 74.57       & 72.22    & 53.53     \\
RoBERTa & zh ori: OCNLI           & 64.49       & \cellcolor{Gray}{71.81}   & 73.66   & 52.95 & 51.8  & 73.43  & 73.86     & 71.79    & 54.16     \\
RoBERTa & zh: OCNLI + XNLI        & 71.01 & 59.39   & 99.06   & 55.87 & 54.64 & 76.83  & 76.50      & 75.48    & \cellcolor{Gray}{70.18}     \\ \hdashline %
XLM-R   & zh MT: XNLI             & 66.87 & 55.53   & \cellcolor{Gray}{\textbf{99.96}}   & 56.11 & 55.29 & 77.69  & 77.9   & 74.37    & 46.81     \\
XLM-R   & zh ori: OCNLI           & 69.08 & 71.29   & 88.63   & 55.93 & 55.05 & 76.84  & 77.00   & 71.42    & 65.51     \\
XLM-R   & zh: OCNLI + XNLI        & \cellcolor{Gray}{71.49} & 61.85   & 99.45   & \cellcolor{Gray}{58.15} & \cellcolor{Gray}{57.92} & \cellcolor{Gray}{79.16}  & \cellcolor{Gray}{79.28}     & \cellcolor{Gray}{\textbf{77.93}}    & 61.88     \\ \midrule %
XLM-R   & en:MNLI                 & 67.94 & 65.77   & \cellcolor{Gray}{99.2}    & \cellcolor{Gray}{55.14} & \cellcolor{Gray}{54.6}  & \cellcolor{Gray}{75.75}  & \cellcolor{Gray}{75.76}    & 70.76    & 50.90      \\
XLM-R   & en: En-all-NLI          & \cellcolor{Gray}{74.52}       & \cellcolor{Gray}{80.36}   & 97.58   &54.74 & 53.56 & 73.96  & 73.92    & \cellcolor{Gray}{71.02}    & \cellcolor{Gray}{82.73}     \\ \midrule %
XLM-R   & mix: OCNLI + En-all-NLI & \cellcolor{Gray}{\textbf{77.36}} & \cellcolor{Gray}{\textbf{81.93}}   & 95.09   & \cellcolor{Gray}{\textbf{59.23}} & \cellcolor{Gray}{\textbf{58.00}}    & \cellcolor{Gray}{\textbf{79.88}}  & \cellcolor{Gray}{\textbf{79.92}}  & 74.53    & \cellcolor{Gray}{\textbf{87.77}}     \\
XLM-R   & mix: XNLI + En-all-NLI  & 73.57 & 66.15   & \cellcolor{Gray}{99.68}   & 57.02 & 55.51 & 78.38  & 78.53    & \cellcolor{Gray}{75.15}    & 80.33\\\midrule
Human & &85.00 &  85.00 & 98.00 & 83.00 & 83.00 & 83.00 &  83.00 & 78.00 & 98.00 \\
\bottomrule   
\end{tabular}
}
\caption{Accuracy on the stress test. Distr H/P(-n): distraction in Hypothesis/Premise (with negation).
    \label{tab:stress}}
\end{table*}

\begin{table*}[t]
\centering
\footnotesize
\scalebox{.75}{
\begin{tabular}{ll|c|ccccc|ccccccccc}
\toprule
Model &  Fine-tuned on & Overall  
& \rotatebox{90}{\parbox{1.5cm}{$\ast$ Classifier}} 
& \rotatebox{90}{\parbox{1.5cm}{$\ast$ Idioms}} 
& \rotatebox{90}{\parbox{1.5cm}{$\ast$ Non-core\\arguement}}
& \rotatebox{90}{\parbox{1.5cm}{$\ast$ Pro-drop}}
& \rotatebox{90}{\parbox{1.5cm}{$\ast$ Time \\of event}}
& \rotatebox{90}{\parbox{1.5cm}{Anaphora}} 
& \rotatebox{90}{\parbox{1.5cm}{Argument \\structure}}  
& \rotatebox{90}{\parbox{1.5cm}{Common \\sense}} 
& \rotatebox{90}{\parbox{1.5cm}{Comparatives}}
& \rotatebox{90}{\parbox{1.5cm}{Double \\negation}}
& \rotatebox{90}{\parbox{1.5cm}{Lexical \\semantics}}
& \rotatebox{90}{\parbox{1.5cm}{Monotonicity}} 
& \rotatebox{90}{\parbox{1.5cm}{Negation}} 
& \rotatebox{90}{\parbox{1.5cm}{World \\knowledge}} \\\midrule
RoBERTa & zh MT: XNLI-small & 62.9 & 65.8 & 64.7 & 55.2 & 80.5 & 60.0 & 59.6 & 67.4 & 54.3 & 61.4 & 48.3 & 60.9 & 59.7 & 66.2 & 39.0 \\
RoBERTa & zh MT: XNLI & 67.7 & {67.6} & 66.2 & {59.4} & 82.3 & 65.1 & 69.9 & 72.0 & \cellcolor{Gray}{56.8} & 70.4 & 64.2 & 67.5 & 61.7 & 72.9 & 52.1 \\
RoBERTa & zh ori: OCNLI & 67.8 & 62.0 & {68.0} & \cellcolor{Gray}{59.4} & 80.7 & {77.5} & {70.3} & 70.0 & 56.0 & 66.6 & 64.2 & 68.4 & 61.7 & 72.4 & 57.9 \\
RoBERTa & zh: OCNLI + XNLI & \ul{69.3} & 66.3 & 67.1 & 58.6 & {83.0} & 74.0 & 70.1 & {73.5} & 54.9 & {74.1} & {67.5} & {69.1} & {62.5} & {76.0} & {60.0} \\ \hdashline %
XLM-R & zh MT: XNLI & 60.9 & 61.2 & 62.3 & 50.4 & 71.9 & 59.7 & 60.3 & 63.3 & 51.7 & 65.2 & 54.9 & 61.0 & 53.5 & 66.9 & 58.3 \\
XLM-R & zh ori: OCNLI & 68.0 & 57.6 & 70.1 & 58.0 & 79.6 & 76.3 & 67.4 & 70.3 & 55.3 & 69.8 & 75.8 & 71.1 & 62.5 & 71.1 & 62.1 \\
XLM-R & zh: OCNLI + XNLI & \cellcolor{Gray}{71.5} & \cellcolor{Gray}{70.4} & \cellcolor{Gray}{71.6} & 57.5 & \cellcolor{Gray}{84.6} & \cellcolor{Gray}{77.8} & \cellcolor{Gray}{74.5} & \cellcolor{Gray}{74.7} & 55.3 & \cellcolor{Gray}{75.5} & \cellcolor{Gray}{76.7} & \cellcolor{Gray}{72.8} & \cellcolor{Gray}{62.7} & \cellcolor{Gray}{76.3} & \cellcolor{Gray}{65.3} \\ \midrule %
XLM-R & en: MNLI & 70.2 & 70.1 & 73.9 & \cellcolor{Gray}{57.5} & 86.4 & 70.8 & 69.3 & 72.9 & 48.9 & 76.0 & 62.5 & 67.8 & 62.6 & 77.0 & 62.1 \\
XLM-R & en: En-all-NLI & \cellcolor{Gray}{71.9} & \cellcolor{Gray}{71.8} & \cellcolor{Gray}{\textbf{74.3}} & 56.2 & \cellcolor{Gray}{87.4} & \cellcolor{Gray}{75.7} & \cellcolor{Gray}{74.9} & \cellcolor{Gray}{74.8} & \cellcolor{Gray}{49.1} & \cellcolor{Gray}{80.5} & \cellcolor{Gray}{70.8} & \cellcolor{Gray}{69.1} & \cellcolor{Gray}{63.8} & \cellcolor{Gray}{77.8} & \cellcolor{Gray}{64.2} \\ \midrule %
XLM-R & mix: OCNLI + En-all-NLI & \cellcolor{Gray}{\textbf{74.9}} & \cellcolor{Gray}{\textbf{72.7}} & \cellcolor{Gray}{\textbf{74.3}} & 60.1 & \cellcolor{Gray}{\textbf{88.5}} & \cellcolor{Gray}{\textbf{84.5}} & \cellcolor{Gray}{\textbf{77.3}} & \cellcolor{Gray}{\textbf{78.1}} & 56.6 & \cellcolor{Gray}{\textbf{81.3}} & \cellcolor{Gray}{\textbf{79.2}} & \cellcolor{Gray}{\textbf{77.2}} & 65.6 & \cellcolor{Gray}{\textbf{78.0}} & 67.9 \\
XLM-R & mix: XNLI + En-all-NLI & 71.4	&70.2	&58.5	& \cellcolor{Gray}{\textbf{85.5}}	&71.3	&75.2	&75.5	&55.1	& \cellcolor{Gray}{\textbf{79.2}}	&70.0	&69.1	&62.4	& \cellcolor{Gray}{\textbf{76.2}}	&72.1	& \cellcolor{Gray}{\textbf{71.3}} \\\bottomrule
\end{tabular}
}
\caption{Accuracy on the expanded diagnostics. Uniquely Chinese linguistic features
    are prefixed with $\ast$. 
    \label{tab:diagnostics}}
\end{table*}

\begin{table*}[t]
\footnotesize
\centering
\scalebox{.85}{
\begin{tabular}{ll|c|cccccc}\toprule
	model & finetune on & overall & boolean & comparative & conditional & counting & negation & quantifier \\\midrule
	RoBERTa & zh MT: XNLI-small & 46.57 & 32.81 & 34.41 & 61.48 & 81.82 & 33.27 & 35.63 \\
	RoBERTa & zh MT: XNLI & 50.64 & 33.35 & 39.02 & \mygray{66.55} & 84.51 & 40.92 & 39.50 \\
	RoBERTa & zh ori: OCNLI & 47.53 & 35.81 & 34.81 & 62.87 & 69.64 & 49.84 & 32.24 \\
	RoBERTa & zh: OCNLI + XNLI & 51.13 & 38.16 & 37.98 & 66.19 & 75.73 & 53.31 & 35.43 \\ \hdashline %
	XLM-R & zh ori: OCNLI & \cellcolor{Gray}{54.33} & \cellcolor{Gray}{\textbf{54.19}} & \cellcolor{Gray}{\textbf{49.02}} & 52.46 & 79.70 & \cellcolor{Gray}{59.52} & 31.08 \\
	XLM-R & zh MT: XNLI & 50.79 & 33.39 & 35.33 & 66.01 & 87.23 & 33.17 & \cellcolor{Gray}{\textbf{49.60}} \\
	XLM-R & zh: OCNLI + XNLI & 52.43 & 34.51 & 36.93 & 59.98 & \cellcolor{Gray}{88.70} & 54.37 & 40.08 \\ \midrule %
	XLM-R & en: MNLI & 49.09 & 33.27 & 37.98 & 66.25 & 89.70 & 34.69 & 32.65 \\
	XLM-R & en: En-all-NLI & \cellcolor{Gray}{55.37} & \cellcolor{Gray}{33.43} & \cellcolor{Gray}{39.70} & \cellcolor{Gray}{66.65} & \cellcolor{Gray}{92.34} & \cellcolor{Gray}{64.11} & \cellcolor{Gray}{35.99} \\ \midrule %
	XLM-R & mix: OCNLI + En-all-NLI & \cellcolor{Gray}{\textbf{57.95}} & \cellcolor{Gray}{40.70} & \cellcolor{Gray}{44.49} & 63.67 & 91.54 & \cellcolor{Gray}{\textbf{74.47}} & 32.81 \\
	XLM-R & mix: XNLI + En-all-NLI & 57.73 & 40.30 & 37.82 & \cellcolor{Gray}{\textbf{66.67}} & \cellcolor{Gray}{\textbf{93.19}} & 61.52 & \cellcolor{Gray}{46.87} \\\bottomrule
\end{tabular}
}
\caption{Accuracy on the Chinese semantic probing datasets, designed following 
\citet{probing2020}. 
    \label{tab:probing}}
\end{table*}

\paragraph{Antonyms and Synonyms}

All models except those fine-tuned on OCNLI
achieved almost perfect score on the synonym
test. 
However, for antonyms, both monolingual and multilingual
models fine-tuned with OCNLI perform better
than XNLI. 
XLM-R fine-tuned with English NLI data only 
again outperforms the best of monolingual models ($\sim$80\% vs.~$\sim$72\%). 
Interestingly, adding XNLI to all English NLI
data hurts the accuracy badly (a 14\% drop),
while adding OCNLI to the same English data
improves the result slightly.

As antonyms are harder to learn \citep{breaknli}, we take our results
to mean that either expert-annotated data for Chinese or a huge English NLI dataset
is needed for a model to learn decent 
representations about antonyms, as indicated by the high performance of RoBERTa fine-tuned with OCNLI (71.81\%), and XLM-R fine-tuned with En-all-NLI (80.36\%), on antonyms.  That is,
using machine-translated XNLI will not 
work well for learning antonyms ($\sim$55\% accuracy).

\paragraph{Distraction}

Results in Table~\ref{tab:stress} show that adding distractions to the hypotheses has a more negative
negative impact on models' performance, compared with
appending distractions to the premises. 
The difference is about 20\% for all models (see  ``Distr H'' columns and ``Distr P'' columns in Table~\ref{tab:stress}), 
which has not been reported in previous studies,
to the best of our knowledge. 
Including a negation in the hypothesis makes it
even more challenging, as we see another
one percent drop in the accuracy for all models. 
This is expected as previous literature
has demonstrated the key role negation
plays when hypotheses are produced by the annotators~\citep{poliak2018hypothesis}.

\paragraph{Spelling}
This is another case where cross-lingual transfer
with English data alone falls behind
monolingual Chinese models (by about 4\%).
Also
the best results are from fine-tuning XLM-R with
OCNLI + XNLI, rather than a combination
of English and Chinese data. 
Considering the data is created by swapping
Chinese characters with others 
of the same pronunciation, we take it
to suggest that monolingual models
are still better at picking up
the misspellings or learning the connections
between characters at 
the phonological level.

\paragraph{Numerical Reasoning}
Results in the last column of Table~\ref{tab:stress} suggest a similar pattern: 
using all English NLI data for cross-lingual 
transfer outperforms the best monolingual model.
However, fine-tuning
a monolingual model with the small OCNLI (50k examples, accuracy: 54\%) achieves better accuracy
than using a much larger MNLI (390k examples, accuracy: 51\%) for cross-lingual
transfer, although both are worse than XLM-R fine-tuned with all English NLI which has more than 1,000k examples (accuracy: 83\%).
This suggests that there are cases where 
a monolingual setting (RoBERTa with OCNLI) is competitive against 
zero-shot transfer with a large
English dataset (XLM-R with MNLI). However, that competitiveness
may disappear when the English dataset grows
to an order of magnitude larger in size or becomes more diverse (En-all-NLI contains 4 different English NLI datasets).

\subsection{Hand-written diagnostics}

Results on the expanded diagnostics are presented in Table~\ref{tab:diagnostics}. 
We first see that XLM-R fine-tuned with only English
performs very well, at 70.2\% and 71.9\%, even slightly 
higher than the best monolingual Chinese model (69.3\%).  

Most surprisingly, \textbf{in 3/5 categories with uniquely 
Chinese linguistic features,
zero-shot transfer outperforms 
monolingual models}.
Only in ``non-core arguments'' and ``time of event'' do we see
higher performance of OCNLI as the fine-tuning data. 
What is particularly striking is that for ``idioms (\textit{Chengyu})'', XLM-R fine-tuned only on English data achieves the best result, suggesting that
the cross-lingual transfer is capable of learning meaning representation beyond the surface lexical information, at least for many of the idioms we tested. The overall results (accuracy of 74.3\%) indicate that cross-lingual transfer is very successful in most cases. 
Manual inspection of the results shows that for many NLI pairs with idioms, XLM-R correctly predicts the figurative interpretation of the idiom as entailment, and the literal interpretation as non-entailment, as described in section~\ref{sec:new:diagnostics}.  
Looking at OCNLI and XNLI, we observe that 
they perform similarly when fine-tuned on monolingual RoBERTa. \orange{However, when fine-tuned with XLM-R, OCNLI has a clear advantage (68.0\% versus 60.9\%), suggesting that OCNLI may produce more stable results than XNLI.} Furthermore, when coupled
with English data to be used with XLM-R, we see
again that OCNLI + En-all-NLI results in an accuracy 3 percent higher than XNLI + En-all-NLI. 

\subsection{Semantic fragments}

Results on the semantic probing datasets (shown in Table~\ref{tab:probing}) are more mixed.
First, the results are in general much
worse than the other evaluation data,
but overall, XLM-R fine-tuned with OCNLI and
all English data still performs the best.
The overall lower performance is likely due to the longer length of premises and hypotheses in the semantic probing datasets, compared with the other three evaluation sets. 
Second, zero-shot transfer is better
or on par with monolingual Chinese RoBERTa in 4/6 semantic fragments (except Boolean and quantifier). 
Third, for Boolean and comparative,
XLM-R fine-tuned with OCNLI has a much 
better result than all other monolingual models or XLM-R fine-tuned with mixed data.
We also observe that all models have highest performance on the ``counting'' fragment. Note that none of the models have seen any data from the ``counting'' fragment during fine-tuning. That is, all the knowledge come from the pre-training and fine-tuning on general NLI datasets.
The surprisingly good performance of XLM-R model (w/ En-all-NLI, 92.34\%) suggests that it may have already acquired a mapping from counting the words/names to numbers, and this knowledge can be transferred cross-linguistically.

\section{Conclusion and Future Work}

In this paper, we examine the cross-lingual transfer ability of XLM-R in the context of 
Chinese NLI through four new sets of aversarial/probing tasks consisting of a total of 17 new high-quality and linguistically motivated challenge datasets. We find that cross-lingual transfer via fine-tuning solely on benchmark English data generally yields impressive performance. In 3/4 of our task categories, such \emph{zero-shot transfer} outperforms our best monolingual models trained on benchmark Chinese NLI data, including 3/5 of our hand-crafted challenge tasks that test uniquely Chinese linguistic phenomena. These results suggest that multilingual models are indeed capable of considerable cross-lingual linguistic transfer and that zero-shot NLI may serve as a serious alternative to large-scale dataset development for new languages.  

These results come with several important caveats. Model performance is still outperformed by conservative estimates of human performance and our best models still have considerable room for improvement; we hope that our new resources will be useful for continuing to benchmark progress on Chinese NLI. We also find that high-quality Chinese NLI data (e.g., OCNLI) \emph{can} help improve results further, which suggests an important role for certain types of expertly annotated monolingual data in a training pipeline. In virtue of our study being limited to \emph{behavioral testing}, the exact reason for \emph{why} cross-lingual zero-shot transfer generally performs well, especially on some Chinese-specific phenomena, is an open question that requires further investigation. In particular, we believe that techniques that couple behavioral testing with \emph{intervention} techniques \cite{geiger-etal-2020-neural,vig2020causal} and other analysis methods \cite{giulianelli2018under,belinkov2019analysis,sinha2021masked} might provide insight, and that our new resources can play an important role in such future work.

\section*{Acknowledgments}
This research was supported in part by Lilly Endowment, Inc., through its support for the Indiana University Pervasive Technology Institute.
He Zhou is sponsored by China Scholarship Council.

\bibliography{acl2021}
\bibliographystyle{acl_natbib}

\clearpage

\appendix

\section{Details for dataset creation}
\label{sec:appendix:ex}
In this section, we list example NLI pairs and their translations.
For examples of the Chinese HANS and stress tests,
see Table~\ref{tab:ex:hans:stress}.
For the expanded diagnostics, see Table~\ref{app:tab:ex:diagnostics}.
For the semantic/logic probing dataset, see Table~\ref{tab:ex:probing}.

\paragraph{Chinese HANS}
Table \ref{tab:ocnli-heuristics} lists the number of examples in OCNLI for each inference label that satisfy the two heuristics we are examining.
We observe that \textit{entailment} examples take the majority for both heuristics. Therefore, we hypothesize that if the heuristics are learned, the \textit{entailment} examples are likely to be correctly predicted while \textit{non-entailment} (contradiction and neutral) examples are prone to receive wrong prediction.

To guarantee the generated sentences are syntactically and semantically sound, we add features for our vocabulary so that subject- predicate and verb-object constraints are satisfied, e.g., some verbs can only take animate subjects and objects.  
We then generate 50 premise-hypothesis pairs for each template described in our Github repository.\footnote{\urlstyle{rm}\url{https://github.com/huhailinguist/ChineseNLIProbing}} Excluding duplicated examples, our generated dataset has 1,941 pairs and the distribution of the three labels is shown in Table \ref{tab:hans-generate-data}. 

The templates for the two heuristics are listed in Appendix~\ref{sec:app:hans:templates}.

\begin{table}[t]
\centering
\scalebox{0.8}{
\begin{tabular}{cccc}
\midrule
Heuristic       & entailment & contradiction & neutral \\\midrule
lexical overlap & 944        & 155           & 109     \\
subsequence     & 190       & 10            & 18     \\
\midrule
\end{tabular}}
\caption{Distribution of the two heuristics in OCNLI.}
\label{tab:ocnli-heuristics}
\end{table}

\begin{table}[t]
\centering
\scalebox{0.8}{
\begin{tabular}{cccc}
\midrule
Heuristic       & entailment & contradiction & neutral \\\midrule
lexical overlap & 441        & 647           & 340     \\
subsequence     & 100        & 193            & 220     \\\midrule
Total  & 541   &  840 &  560 \\\midrule
\end{tabular}}
\caption{Distributional statistics of our synthesized Chinese HANS dataset. }
\label{tab:hans-generate-data}
\end{table}

\paragraph{Antonym}
After looking at the quality of initially generated data,
we decided to replace only the nouns and adjectives with their antonyms since such replacements are most likely result in grammatical and contradictory hypotheses.\footnote{We use the LTP toolkit \urlstyle{rm} (\url{https://github.com/HIT-SCIR/ltp}) to annotate the POS tags and our antonym list is from  \url{https://github.com/liuhuanyong/ChineseAntiword}.}

\paragraph{Synonym}
After inspecting the initially generated data,
we decided to perform replacements only to verbs and adjectives. 
To ensure the quality of synonyms, we rank the synonyms from a commonly used
synonym dictionary by their vector similarity to the original word,
and pick the top ranking synonym.\footnote{We use the synonym list
from \urlstyle{rm} \url{https://github.com/Keson96/SynoCN} and the similarity score
from the Python package Synonyms at \url{https://github.com/chatopera/Synonyms}.}

\paragraph{Distraction}
We created the distraction data similar to the stress test setting~\citep{naik2018stress} but experimented with variations as to where ``distractor statement''---either a tautology or a true statement---was added: the premise or the hypothesis. The distractor statement also varied w.r.t.~whether it contains a negation:

\begin{itemize}
\setlength{\itemsep}{0pt}
    \setlength{\parskip}{0pt}
    \setlength{\parsep}{0pt}
\item \textbf{Premise-no-negation}: A distractor statement is added to the end of the premise and it contains no negation.

\item \textbf{Premise-negation}: A distractor statement containing a negation is added to the premise. 

\item \textbf{Hypothesis-no-negation}: A distractor statement is added to the end of the hypothesis. 

\item \textbf{Hypothesis-negation}: Same as the previous condition except that the distractor contains a negation. 

\end{itemize}

 Only two tautologies are used in \citet{naik2018stress}. In this paper, to thoroughly examine the influence of different true statements, we designed 50 tautology/statements varied in three factors: length, out-of-vocabulary, and negation word. There are 25 statements pairs in total (1 tautology and 24 true statements); each pair includes a true statement and its corresponding true statement with negation form. All the statements range from 5 to 16 characters. For the true statements in negation form, two common Chinese negation words 不~and  没~are used for negation. For the 24 true statement pairs, half of them contains at least one Out-of-Vocabulary word from OCNLI.\footnote{See \urlstyle{rm}\url{https://github.com/huhailinguist/ChineseNLIProbing} for details.}

Experiments show that length, Out-of-Vocabulary words, and the choice of negator have little effects on the results.

\paragraph{Spelling} 
We generate a set of data containing ``spelling errors" by replacing one random character in the hypotheses with its homonym, which is defined as a character with the same \textit{pinyin} pronunciation ignoring tones. We also limit the frequency of the homonym as within the range of 100 to 6000 so that the character is neither too rare nor too frequent. 
 
\paragraph{Numerical reasoning} We extracted sentences from Ape210k \citep{ape210k}, a large-scale math word problem dataset containing 210K Chinese elementary school-level problems\footnote{We split all problems into individual sentences and filter out sentences without numbers. Then we remove sentences without any named entities (``PERSON", ``LOCATION" and ``ORGANIZATION") using the NER tool provided by LTP toolkit \citep{LTP4che2020n}.}.  We generate entailed, contradictory and neutral hypotheses for each premise, with the rules below:
\begin{enumerate}
      \setlength{\itemsep}{0pt}
    \setlength{\parskip}{0pt}
    \setlength{\parsep}{0pt}
    \item \textbf{Entailment:} Randomly choose a number x and change it to y from the hypothesis. If the y $>$ x, prefix it with one phrase that translate to ``less than"; if y $<$ x, prefix it with one phrase that translate to ``more than".\\
    Premise: \textit{Mary types 110 words per minute.} Hypothesis: \textit{Mary types less than 510 words per minute.} 
    \item \textbf{Contradiction:} Perform either 1) randomly choose a number x from the hypothesis and change it; 2) randomly choose a number from the hypothesis and prefix it with one phrase that translate to ``less than" or ``more than".\\
    Premise: \textit{Mary types 110 words per minute.} Hypothesis: \textit{Mary types 710 words per minute.} 
    \item \textbf{Neutral:} Reverse the corresponding entailed premise-hypothesis pairs.\\
    Premise: \textit{Mary types less than 510 words per minute.}  Hypothesis: \textit{Mary types 110 words per minute.}
\end{enumerate}
The result contains 2,871 unique premise sentences and 8,613 NLI pairs.

\paragraph{Diagnostics}

The diagnstics for \textit{classifiers} (or measure word) and \textit{non-core arguments} 
are explained in detail below (see examples in Table~\ref{app:tab:ex:diagnostics}). 

\begin{enumerate}
      \setlength{\itemsep}{0pt}
    \setlength{\parskip}{0pt}
    \setlength{\parsep}{0pt}
        \item \textit{classifiers} (or measure word): in Chinese, when modified by a numeral, a noun must be preceeded by a category of words called classifier. They can be semantically vacuous but sometimes also carry semantic content:  一\underline{匹}狼~\textit{one \ul{pi} wolf} (one wolf); 一\underline{群}狼~\textit{one \ul{qun} wolf} (one \ul{pack} of wolves). Our examples require the model to understand the semantic content of the classifiers. 
        \item \textit{non-core arguments}: in Chinese syntax, sometimes a noun phrase at the argument position (e.g., object) is not serving as an object: 今天吃筷子，不吃叉子。\textit{today eat chopsticks, not eat fork} (We eat \textbf{with} chopsticks today, not with fork). \citet{noncoreargs} shows that this structure is very productive in Chinese and we take example sentences from her dissertation. 
\end{enumerate}

Additionally, for the \textit{pro-drop} examples, they are constructed such that the models return the correct inference relation only when they successfully identify what the dropped \textit{pro} refers to. That is, our constructed premises involve several entities the dropped \textit{pro} could potentially refer to, and the entailed hypothesis identifies the correct referent while the neutral/contradictory hypothesis does not
(see Table~\ref{app:tab:ex:diagnostics} for an example).

\begin{table*}[t]
\footnotesize
\begin{adjustbox}{max width=\textwidth}
\begin{tabular}{p{3cm}p{.5cm}p{5cm}p{5cm}l}\toprule
category & n & premise & hypothesis & label \\\midrule
CLUE \citep{clue} & 514 & 有些学生喜欢在公共澡堂里唱歌。\newline Some students like to sing in public showers. & 有些女生喜欢在公共澡堂里唱歌。\newline Some female students like to sing in public showers. & N \\\hdashline
CLUE expansion (ours) & 800 & 雷克雅未克所有旅馆的床位加在一起才一千六百个。\newline There are only one thousand six hundred beds in all hotels in Reykjavik. & 雷克雅未克有旅馆的床位超过一千个。\newline  Some hotel in Reykjavik has over a thousand beds. & N \\\hdashline
World Knowledge (ours) & 37 & 上海在北京的南边。\newline Shanghai is to the south of Beijing. & 北京在上海的南边。\newline Beijing is to the south of Shanghai. & C \\\hdashline
Classifier (ours) & 138 & 这些孩子吃了一个苹果。\newline These children ate an apple. & 这些孩子吃了一筐苹果。\newline These children ate a basket of apples. & N \\\hdashline
Chengyu/idioms (ours) & 250 & 这帮人可狡猾得很啊，你一个电话打过去，\underline{打草惊蛇}，后果不堪设想。\newline These people are so cunning! If you call them, you would \underline{hit grass alert snake}. The consequences would be unimaginable. & 你打电话过去会让这帮人察觉，造成不好的结果。\newline If you call them, it will alert them, and bring negative consequences. & E \\
& & \textit{same as above} & 这些狡猾的人养了很多蛇。\newline These cunning people have raised a lot of snakes. & N\\\hdashline
Pro-drop (ours) & 197 & 见了很多学生，又给老师们开了两个小时会，校长和主任终于可以下班了。\newline After (\textit{pro}) meeting many students and (\textit{pro}) having two hours of meeting with the teachers, the principal and the director can finally get off work. & 校长见了很多学生。\newline The principal met many students. & E \\
& & \textit{same as above} & 老师们见了很多学生。\newline The teachers met many students. & N\\\hdashline
Non-core arguments (ours) & 185 & 平时范志毅都\underline{踢后卫}的，今天却改当前锋了。\newline Zhiyi Fan usually \underline{kicks full back} (meaning ``plays full back in soccer games''), but today he switched to playing forward. & 范志毅经常用腿踢对方的后卫。\newline Zhiyi Fan usually uses his legs to kick the other team's full back. & N \\\bottomrule
\end{tabular}
\end{adjustbox}
\caption{Example NLI pairs in expanded diagnostics with translations. \label{app:tab:ex:diagnostics}}
\end{table*}
\end{CJK*}

\section{Hyperparameters for experiments}
\label{sec:app:finetune}

Table~\ref{tab:parameter} presents the hyperparameters used for the models. The learning-rate search space for RoBERTa is: 1e-5, 2e-5, 3e-5, 4e-5 and 5e-5, for XLM-R: 5e-6, 7e-6, 9e-6, 2e-5 and 5e-5. 

\begin{table}[t]
\small
\centering
\scalebox{1}{
\begin{tabular}{lll}
\toprule
Model   & Training Data           & LR \\ \midrule
RoBERTa & zh MT: XNLI-small       &  3e-05      \\
RoBERTa & zh MT: XNLI             &  2e-05      \\
RoBERTa & zh ori: OCNLI           &  2e-05      \\
RoBERTa & zh: OCNLI + XNLI        &  3e-05      \\ \hdashline
XLM-R   & zh MT: XNLI             &  7e-06      \\
XLM-R   & zh ori: OCNLI           &  5e-06      \\
XLM-R   & zh: OCNLI + XNLI        &  7e-06      \\ \midrule
XLM-R   & en:MNLI                 &  5e-06      \\
XLM-R   & en: En-all-NLI          &  7e-06      \\ \midrule
XLM-R   & mix: OCNLI + En-all-NLI &  7e-06      \\
XLM-R   & mix: XNLI + En-all-NLI  &  7e-06    \\ \bottomrule 
\end{tabular}
}
\caption{Hyper-parameters used for fine-tuning the models. All models are fine-tuned for 3 epochs with maximum length of 128.  \label{tab:parameter}}
\end{table}

\section{Chinese-to-English transfer}
\label{sec:app:ch2en:transfer}

We present Chinese-to-English transfer results in this section.  
As mentioned in the main text, for most of the
cases, zero-shot transfer learning does not work well
mostly likely due to the small size of 
OCNLI. However, for 3 out of the 4 datasets, XLM-R fine-tuned with the mix data outperforms the monolingual setting, suggesting that even OCNLI is only 1/20 of En-all-NLI, XLM-R can still acquire some useful information from OCNLI, in addition to what is present in En-all-NLI.  

Specifically, (1) for English HANS, XLM-R fine-tuned with OCNLI is about 13 percentage points below the best
English monolingual model, shown in Table~\ref{tab:app:en-hans}. 
(2) For stress tests shown in Table~\ref{tab:app:en-stress}, the gap is about 5 percent (XLM-R with OCNLI = 74\%; RoBERTa with En-all-NLI = 79\%). Interestingly, XLM-R with OCNLI performs the best for Negation and Word overlap. It even outperforms RoBERTa w/ MNLI on the Antonym, which seems to be consistent with the high performance of OCNLI-trained models on the Chinese Antonym in our constructed stress tests. 
(3) For semantic probing data, as shown in Table~\ref{tab:app:en:probing},
XLM-R with OCNLI is 5 percent behind monolingual model fine-tuned with all English NLI, but performs better than the monolingual RoBERTa fine-tuned with MNLI (53.6\% vs.~51.3\%). This is quite surprising since the size of OCNLI is only 1/8 of MNLI. 
(4) For the English diagnostics as shown in Table~\ref{tab:app:en-diagnostics-i} and Table~\ref{tab:app:en-diagnostics-ii}, XLM-R with OCNLI
is 7 percent behind RoBERTa fine-tuned with MNLI.

We leave it for future work to thoroughly examine transfer learning from a ``low-resource'' language such as Chinese to the high-resource one such as English. 

\begin{table*}[t]
\centering
\begin{adjustbox}{max width=.9\textwidth}
\begin{tabular}{llcccccc}\toprule
Model & Fine-tuned on & Overall & Lexical\_overlap & Subsequence & Constituent & Entailment & Non-entailment \\ \midrule
RoBERTa & en: En-all-NLI & 76.54 & \mygray{96.79} & 67.77 & 65.06 & 99.81 & 53.27 \\
RoBERTa & en: MNLI & \mygray{77.63} & 95.60 & \mygray{\textbf{68.08}} & \mygray{69.21} & 99.74 & \mygray{55.52} \\
\hdashline
XLM-R & en: En-all-NLI & 75.72 & 95.52 & 62.99 & 68.63 & \mygray{99.91} & 51.52 \\
XLM-R & en: MNLI & 74.80 & 92.92 & 65.24 & 66.23 & 98.83 & 50.76 \\ \midrule
XLM-R & zh ori: OCNLI & 64.37 & 71.28 & 54.42 & 67.41 & 98.39 & 30.35 \\
XLM-R & zh MT: XNLI & 68.83 & 81.67 & \mygray{62.07} & 62.74 & \mygray{99.13} & 38.53 \\
XLM-R & zh mix: OCNLI+XNLI & \mygray{71.30} & \mygray{82.52} & 61.72 & 69.66 & 99.08 & \mygray{43.52} \\ \midrule
XLM-R & mix: OCNLI+En-all-NLI & \mygray{\textbf{78.56}} & \mygray{\textbf{96.92}} & \mygray{64.91} & \mygray{\textbf{73.84}} & 99.92 & \mygray{\textbf{57.20}} \\
XLM-R & mix: XNLI+En-all-NLI & 74.65 & 93.93 & 60.97 & 69.04 & \mygray{\textbf{99.96}} & 49.34 \\ \bottomrule
\end{tabular}
\end{adjustbox}
\caption{Results of English HANS \cite{hans}.}
\label{tab:app:en-hans}
\end{table*}

\begin{table*}[t]
\centering
\begin{adjustbox}{max width=\textwidth}
\begin{tabular}{llcccccccccc}
\midrule
Model & Fine-tuned on & Overall & Antonym & \parbox{1.5cm}{\centering Content\\word swap} & \parbox{1.5cm}{\centering Function\\word swap} & Keyboard & Swap & \parbox{1.5cm}{\centering Length\\mismatch} & Negation & \parbox{1.5cm}{\centering Numerical\\reasoning} & \parbox{1.5cm}{\centering Word\\overlap} \\ \hline
RoBERTa & en: En-all-NLI & 79.48 & 82.91 & \mygray{\textbf{86.22}} & 88.71 & \mygray{\textbf{87.8}} & \mygray{\textbf{87.48}} & \mygray{\textbf{88.28}} & 60.25 & 79.26 & 62.85 \\
RoBERTa & en: MNLI & 77.9 & 69.03 & 85.74 & \mygray{\textbf{88.75}} & 87.39 & 87.05 & 88.23 & 59.19 & 65.46 & 61.48 \\\hdashline
 XLM-R & en: En-all-NLI & \mygray{79.6} & \mygray{86.25} & 85.26 & 87.38 & 86.31 & 86.72 & 87.25 & \mygray{61.06} & \mygray{\textbf{82.84}} & \mygray{65.79} \\
 XLM-R & en: MNLI & 77.6 & 74.65 & 85.09 & 87.33 & 86.08 & 86.42 & 86.96 & 60.95 & 54.66 & 65.13 \\\hline
 XLM-R & zh ori: OCNLI & 74.31 & \mygray{72.52} & 75.12 & 77.71 & 76.27 & 76.39 & 77.23 & \mygray{\textbf{72.86}} & \mygray{55.85} & \mygray{\textbf{72.79}} \\
 XLM-R & zh MT: XNLI & 77.78 & 65.12 & \mygray{85.11} & 86.64 & 85.79 & 85.71 & 85.91 & 63.52 & 43.95 & 71.63 \\
 XLM-R & zh mix: OCNLI+XNLI & \mygray{77.83} & 66.83 & 84.96 & \mygray{86.69} & \mygray{85.81} & \mygray{85.87} & \mygray{85.98} & 63.97 & 51.56 & 68.38 \\\hline
 XLM-R & mix: OCNLI+En-all-NLI & \mygray{\textbf{80.01}} & \mygray{\textbf{86.33}} & 85.22 & \mygray{87.4} & 86.26 & \mygray{86.77} & \mygray{87.23} & \mygray{62.52} & \mygray{81.79} & \mygray{67.54} \\
 XLM-R & mix: XNLI+En-all-NLI & 79.38 & 85.27 & \mygray{85.35} & 87.2 & \mygray{86.28} & 86.74 & 87.22 & 60.29 & 80.5 & 66.19 \\
\bottomrule

\end{tabular}
\end{adjustbox}
\caption{Results of English stress test \cite{naik2018stress}.}
\label{tab:app:en-stress}
\end{table*}

\begin{table*}[t]
\centering
\begin{adjustbox}{max width=\textwidth}
\begin{tabular}{llccccccccc}
\toprule
Model & Fine-tuned on & Overall & Boolean & Comparative & Conditional & Counting & \parbox{1.5cm}{\centering Monotonicity\\hard} & \parbox{1.5cm}{\centering Monotonicity\\simple} & Negation & Quantifier \\ \midrule
RoBERTa & en: MNLI & 51.31 & 43.58 & 39.6 & 66.24 & 63.34 & 61.28 & 60.1 & 37.26 & 39.08 \\
RoBERTa & en: En-all-NLI & 58.72 & 60.18 & 40.28 & \mygray{\textbf{66.3}} & \mygray{66.22} & 59.6 & 58.98 & 64.46 & \mygray{\textbf{53.74}} \\\hdashline
 XLM-R & en: MNLI & 53.54 & 59.16 & 41.62 & 66.3 & 61.72 & 63.26 & \mygray{\textbf{62.82}} & 33.52 & 39.92 \\
 XLM-R & en: En-all-NLI & \mygray{59.85} & \mygray{\textbf{71.58}} & \mygray{45.18} & 66.3 & 60.4 & \mygray{63.86} & 62.02 & \mygray{65.68} & 43.78 \\\hline
 XLM-R & zh ori: OCNLI & 53.61 & \mygray{66.02} & \mygray{\textbf{60.62}} & 41.1 & 58.0 & 47.86 & 49.88 & \mygray{51.88} & \mygray{53.5} \\
 XLM-R & zh MT: XNLI & 52.29 & 43.24 & 39.0 & 66.22 & 65.66 & 58.08 & \mygray{62.74} & 34.12 & 49.24 \\
 XLM-R & zh mix: OCNLI+XNLI & \mygray{54.68} & 54.64 & 38.84 & \mygray{66.28} & \mygray{\textbf{67.38}} & \mygray{58.18} & 61.38 & 41.88 & 48.82 \\\hline
 XLM-R & mix: OCNLI+En-all-NLI & \mygray{\textbf{60.2}} & \mygray{71.2} & 42.58 & \mygray{66.3} & 62.4 & \mygray{\textbf{64.72}} & 60.9 & \mygray{\textbf{68.58}} & \mygray{44.88} \\
 XLM-R & mix: XNLI+En-all-NLI & 60.06 & 65.8 & \mygray{46.86} & 66.3 & \mygray{65.54} & 61.56 & \mygray{61.44} & 68.5 & 44.48 \\
\bottomrule

\end{tabular}
\end{adjustbox}
\caption{Results of English semantic probing datasets \citep{probing2020}. }
\label{tab:app:en:probing}
\end{table*}

\begin{table*}[t]
\centering
\begin{adjustbox}{max width=\textwidth}
\begin{tabular}{llccccccccccccccccc}
\hline
Model & Fine-tuned on & Overall & \rotatebox{90}{\parbox{1.5cm}{active\\passive}} & \rotatebox{90}{\parbox{1.5cm}{anaphora\\coreference}} & \rotatebox{90}{\parbox{1.5cm}{common sense}} & \rotatebox{90}{\parbox{1.5cm}{conditionals}} & \rotatebox{90}{\parbox{1.5cm}{anaphora\\coreference}} & \rotatebox{90}{\parbox{1.5cm}{coordination scope }} & \rotatebox{90}{\parbox{1.5cm}{core args}} & \rotatebox{90}{\parbox{1.5cm}{datives}} & \rotatebox{90}{\parbox{1.5cm}{disjunction}} & \rotatebox{90}{\parbox{1.5cm}{double negation}} & \rotatebox{90}{\parbox{1.5cm}{downward monotone}} & \rotatebox{90}{\parbox{1.5cm}{ellipsis\\implicits}} & \rotatebox{90}{\parbox{1.5cm}{existential}} & \rotatebox{90}{\parbox{1.5cm}{factivity}} & \rotatebox{90}{\parbox{1.5cm}{genitives\\partitives}} & \rotatebox{90}{\parbox{2cm}{intersectivity}} \\ \hline
RoBERTa & en: MNLI & 66.87 & \mygray{\textbf{62.35}} & 67.59 & \mygray{\textbf{69.47}} & 62.5 & 78.0 & 63.5 & 69.62 & \mygray{\textbf{85.0}} & 39.47 & \mygray{\textbf{92.86}} & \mygray{\textbf{19.33}} & 65.29 & 65.0 & 62.06 & \mygray{\textbf{95.0}} & 60.43 \\
RoBERTa & en: En-all-NLI & \mygray{\textbf{68.03}} & 61.76 & \mygray{\textbf{70.0}} & 69.33 & \mygray{\textbf{63.75}} & \mygray{\textbf{82.5}} & \mygray{\textbf{68.0}} & \mygray{\textbf{75.77}} & 85.0 & \mygray{\textbf{41.58}} & 92.14 & 18.67 & \mygray{\textbf{67.65}} & 65.0 & \mygray{\textbf{62.35}} & 94.0 & 59.57 \\\hdashline
 XLM-R & en: MNLI & 63.03 & 61.76 & 62.76 & 59.73 & 55.62 & 76.0 & 61.5 & 61.54 & 85.0 & 26.84 & 91.43 & 16.0 & 64.12 & 69.0 & 51.47 & 90.0 & 60.0 \\
 XLM-R & en: En-all-NLI & 64.57 & 61.76 & 65.17 & 61.47 & 60.0 & 76.0 & 66.0 & 65.77 & 85.0 & 33.16 & 89.29 & 14.0 & 62.94 & \mygray{\textbf{71.0}} & 58.53 & 90.0 & \mygray{\textbf{60.87}} \\\hline
 XLM-R & zh ori: OCNLI & 59.67 & 60.0 & 59.31 & 57.2 & 58.12 & 70.0 & 56.5 & \mygray{61.54} & \mygray{85.0} & 30.0 & 67.14 & \mygray{17.33} & 54.71 & \mygray{66.0} & 46.18 & \mygray{90.0} & \mygray{59.57} \\
 XLM-R & zh MT: XNLI & 61.76 & 61.18 & \mygray{64.14} & \mygray{60.67} & \mygray{58.75} & 72.5 & 60.0 & 60.77 & 85.0 & \mygray{33.16} & \mygray{91.43} & 12.67 & \mygray{58.24} & 64.0 & 48.24 & 90.0 & 57.83 \\
 XLM-R & zh mix: OCNLI+XNLI & \mygray{61.78} & \mygray{61.76} & 62.76 & 56.93 & 57.5 & \mygray{74.5} & \mygray{61.0} & 61.54 & 85.0 & 31.05 & 90.0 & 12.0 & 57.65 & 65.0 & \mygray{48.53} & 90.0 & 57.39 \\\hline
 XLM-R & mix: OCNLI+En-all-NLI & \mygray{64.51} & \mygray{61.76} & 63.45 & 61.6 & 58.75 & \mygray{76.0} & \mygray{66.0} & \mygray{67.31} & \mygray{85.0} & \mygray{35.26} & 90.71 & \mygray{15.33} & 60.59 & 68.0 & \mygray{60.59} & 91.0 & \mygray{60.87} \\
 XLM-R & mix: XNLI+En-all-NLI & 64.37 & 61.18 & \mygray{64.83} & \mygray{62.27} & \mygray{61.88} & 73.0 & 65.0 & 65.38 & 85.0 & 35.26 & \mygray{91.43} & 14.67 & \mygray{63.53} & \mygray{70.0} & 57.94 & \mygray{94.0} & 60.87 \\
\bottomrule

\end{tabular}
\end{adjustbox}
\caption{Results of English Diagnostics from GLUE-Part I \cite{glue}.}
\label{tab:app:en-diagnostics-i}
\end{table*}

\begin{table*}[t]
\centering
\begin{adjustbox}{max width=\textwidth}
\begin{tabular}{llccccccccccccccccc}
\hline
Model & Fine-tuned on & \rotatebox{90}{\parbox{1.5cm}{intervals\\numbers}} & \rotatebox{90}{\parbox{1.5cm}{lexical\\entailment }} & \rotatebox{90}{\parbox{1.5cm}{morphological\\negation}} & \rotatebox{90}{\parbox{1.5cm}{named\\entities}} & \rotatebox{90}{\parbox{1.5cm}{negation}} & \rotatebox{90}{\parbox{2.2cm}{nominalization}} & \rotatebox{90}{\parbox{1.5cm}{non-monotone}} & \rotatebox{90}{\parbox{1.5cm}{prepositional\\phrases}} & \rotatebox{90}{\parbox{1.5cm}{quantifiers}} & \rotatebox{90}{\parbox{1.5cm}{redundancy}} & \rotatebox{90}{\parbox{1.5cm}{relative\\clauses}} & \rotatebox{90}{\parbox{1.5cm}{restrictivity}} & \rotatebox{90}{\parbox{1.5cm}{symmetry\\collectivity}} & \rotatebox{90}{\parbox{1.5cm}{temporal}} & \rotatebox{90}{\parbox{1.5cm}{universal}} & \rotatebox{90}{\parbox{1.5cm}{upward\\monotone}} & \rotatebox{90}{\parbox{1.5cm}{world\\knowledge}} \\ \hline
RoBERTa & en: MNLI & 54.74 & 66.71 & \mygray{\textbf{89.23}} & 57.22 & 66.59 & 82.14 & 56.0 & \mygray{86.18} & 78.46 & \mygray{79.23} & \mygray{63.75} & \mygray{55.38} & 67.86 & 56.25 & \mygray{\textbf{84.44}} & \mygray{76.47} & 48.51 \\
RoBERTa & en: En-all-NLI & \mygray{\textbf{63.16}} & \mygray{\textbf{71.57}} & 89.23 & 61.11 & \mygray{\textbf{69.02}} & 84.29 & \mygray{\textbf{57.33}} & 84.41 & 74.62 & 73.85 & 63.75 & 53.08 & \mygray{\textbf{70.0}} & \mygray{\textbf{69.38}} & 83.33 & 73.53 & \mygray{\textbf{49.55}} \\\hdashline
 XLM-R & en: MNLI & 45.79 & 65.57 & 84.62 & 61.11 & 61.95 & 82.14 & 52.0 & 85.88 & \mygray{\textbf{82.69}} & 78.46 & 62.5 & 52.31 & 57.14 & 51.25 & 80.0 & 74.12 & 44.03 \\
 XLM-R & en: En-all-NLI & 45.79 & 69.71 & 84.62 & \mygray{61.67} & 64.15 & \mygray{\textbf{85.71}} & 48.67 & 84.71 & 79.23 & 69.23 & 63.12 & 46.15 & 59.29 & 60.0 & 77.78 & 75.29 & 45.82 \\\hline
 XLM-R & zh ori: OCNLI & 39.47 & 56.29 & 75.38 & 41.11 & 53.41 & 73.57 & 51.33 & 85.59 & 63.08 & 83.08 & \mygray{62.5} & \mygray{\textbf{70.77}} & \mygray{64.29} & \mygray{54.38} & 62.22 & \mygray{\textbf{78.82}} & 47.31 \\
 XLM-R & zh MT: XNLI & 42.11 & \mygray{60.14} & \mygray{84.62} & 61.11 & 61.95 & \mygray{74.29} & \mygray{53.33} & \mygray{\textbf{86.76}} & 73.46 & 84.62 & 60.62 & 60.0 & 57.86 & 40.62 & \mygray{84.44} & 68.24 & 47.31 \\
 XLM-R & zh mix: OCNLI+XNLI & \mygray{43.68} & 59.29 & 83.08 & \mygray{\textbf{63.33}} & \mygray{62.2} & 74.29 & 52.0 & 86.76 & \mygray{76.92} & \mygray{\textbf{85.38}} & 62.5 & 60.77 & 59.29 & 43.75 & 82.22 & 67.65 & \mygray{47.61} \\\hline
 XLM-R & mix: OCNLI+En-all-NLI & \mygray{45.26} & \mygray{69.86} & 85.38 & \mygray{62.22} & \mygray{65.12} & \mygray{85.71} & \mygray{50.67} & \mygray{85.0} & 74.62 & 69.23 & \mygray{\textbf{68.12}} & \mygray{47.69} & \mygray{60.71} & 56.25 & \mygray{77.78} & \mygray{75.29} & 45.67 \\
 XLM-R & mix: XNLI+En-all-NLI & 44.21 & 67.29 & \mygray{86.15} & 62.22 & 63.9 & 83.57 & 49.33 & 84.71 & \mygray{75.0} & \mygray{70.77} & 66.88 & 46.92 & 58.57 & \mygray{57.5} & 74.44 & 74.12 & \mygray{45.82} \\
\bottomrule
\end{tabular}
\end{adjustbox}
\caption{Results of English Diagnostics from GLUE-Part II \cite{glue}.}
\label{tab:app:en-diagnostics-ii}
\end{table*}

\section{Example templates for Chinese HANS}\label{sec:app:hans:templates}

We present the templates for the two heuristics in Chinese HANS in Table~\ref{tab:app:lexical:overlap} and Table~\ref{tab:app:subseq:overlap}.

\renewcommand{\arraystretch}{1.2} %

\begin{CJK*}{UTF8}{gbsn}
\begin{table*}[t]
\footnotesize
\centering
\begin{adjustbox}{max width=\textwidth, max totalheight=\textheight}
\begin{tabular}{p{2.5cm}p{5cm}p{9cm}}
\toprule
Category & Template(Premise $\rightarrow$ Hypothesis) & Example \\\midrule
  Entailment:\newline 被 sentence
  & 
  ${\rm N}_1$ 被 V 在 ${\rm N}_{loc}$ 了。 \newline  $\rightarrow$ ${\rm N}_1$ 在 ${\rm N}_{loc}$。  
  & \begin{tabular}[c]{@{}l@{}}艺术家被关在天文馆了。\\The artist is locked in the planetarium. \\ $\rightarrow$  艺术家在天文馆。\\The artist is inside the planetarium. \end{tabular} \\\hline
  \begin{tabular}[c]{@{}l@{}}Entailment:\\ PP-drop\end{tabular}  & \begin{tabular}[c]{@{}l@{}} ${\rm N}_1$ 在 ${\rm N}_{loc}$ LC V ${\rm N}_2$。 \\  $\rightarrow$ ${\rm N}_1$ V ${\rm N}_2$。  \end{tabular} &  \begin{tabular}[c]{@{}l@{}}领导在咖啡馆附近喝啤酒。\\ The leader is drinking beer near the coffee shop. \\ $\rightarrow$ 领导喝啤酒。\\ The leader is drinking beer. \end{tabular} \\\hline
  \begin{tabular}[c]{@{}l@{}}Entailment:\\ Adverb-连\end{tabular}  & \begin{tabular}[c]{@{}l@{}} 连 V ${\rm N}_1$ 的 ${\rm N}_2$ 都 觉得 ADJ。 \\ $\rightarrow$ ${\rm N}_2$ V ${\rm N}_1$。 \end{tabular} & \begin{tabular}[c]{@{}l@{}}连看话剧的研究生都觉得热闹。\\ Even graduate students watching a drama feel excited. \\ $\rightarrow$ 研究生看话剧。\\ The graduate students are watching a drama. \end{tabular} \\\hline
 \begin{tabular}[c]{@{}l@{}}Entailment:\\ Choice\end{tabular}  & \begin{tabular}[c]{@{}l@{}} ${\rm PN}$ 不是 ${\rm N}_1$, 但是 是 ${\rm N}_2$。 \\ $\rightarrow$ ${\rm PN}$ 是 ${\rm N}_2$。 \end{tabular} & \begin{tabular}[c]{@{}l@{}}她不是医生,但是是科学家。She is not a doctor, but a scientist. \\ $\rightarrow$ 她是科学家。She is a scientist. \end{tabular} \\\hline
   \begin{tabular}[c]{@{}l@{}}Entailment:\\ Adverb-drop\end{tabular}  & \begin{tabular}[c]{@{}l@{}} ${\rm N}_1$ ADV V 过 ${\rm N}_2$。 \\ $\rightarrow$ ${\rm N}_1$ V 过 ${\rm N}_2$ 了。 \end{tabular} & \begin{tabular}[c]{@{}l@{}}法官果然讲过笑话。 As expected, the judge has made jokes. \\ $\rightarrow$ 法官讲过笑话。 The judge has made jokes.\end{tabular} \\\hline
  \begin{tabular}[c]{@{}l@{}}Contradiction: \\ Negation\end{tabular}  & \begin{tabular}[c]{@{}l@{}} ${\rm N}_1$ 没有 ${\rm V}$ 过 ${\rm N}_2$。 \\$\nrightarrow$ ${\rm  N}_1$ ${\rm V}$ 过 ${\rm N}_2$。 \end{tabular} & \begin{tabular}[c]{@{}l@{}}医生没有看过电影。 The doctors has never watched movies. \\ $\nrightarrow$ 医生看过电影。 The doctor has watched movies. \end{tabular} \\\hline
  \begin{tabular}[c]{@{}l@{}}Contradiction: \\ Double Negation\end{tabular}  & \begin{tabular}[c]{@{}l@{}} ${\rm N}_1$ 不是不 ${\rm V}$ ${\rm N}_2$。  \\$\nrightarrow$ ${\rm N}_1$ 不 ${\rm V}$ ${\rm N}_2$。 \end{tabular} & \begin{tabular}[c]{@{}l@{}}清洁工不是不吃午饭。 It's not the case that cleaners do not eat lunch. \\ $\nrightarrow$ 清洁工不吃午饭。 Cleaners do not eat lunch. \end{tabular} \\\hline
  \begin{tabular}[c]{@{}l@{}}Contradiction: \\ Swap\end{tabular}  & \begin{tabular}[c]{@{}l@{}} ${\rm PN}$ 把 ${\rm N}_1$ ${\rm V}$ 在 ${\rm N}_{loc}$ 了。  \\$\nrightarrow$ ${\rm N}_1$ 把 ${\rm PN}$ ${\rm V}$ 在 ${\rm N}_{loc}$ 了。 \end{tabular} & \begin{tabular}[c]{@{}l@{}}我们把银行职员留在电影院了。\\ We left the bank clerk in the cinema. \\ $\nrightarrow$ 银行职员把我们留在电影院了。\\ The bank clerk left us in the cinema. \end{tabular} \\\hline
 \begin{tabular}[c]{@{}l@{}}Contradiction: \\ Choice\end{tabular}  & \begin{tabular}[c]{@{}l@{}} ${\rm N}_1$ 本来 想 ${\rm V}_1$ ${\rm N}_2$, 结果 ${\rm V}_2$ ${\rm N}_3$ 了。 \\$\nrightarrow$ ${\rm N}_1$ 本来 想 ${\rm V}_2$ ${\rm N}_3$。 \end{tabular} & \begin{tabular}[c]{@{}l@{}}教授本来想喝啤酒,结果吃西瓜了。\\ The professor was thinking to drink beer but ate \\ watermelon instead. \\ $\nrightarrow$ 教授本来想吃西瓜。\\ The professor was thinking to eat watermelon. \end{tabular} \\\hline
\begin{tabular}[c]{@{}l@{}}Contradiction: \\ Condition\end{tabular}  & \begin{tabular}[c]{@{}l@{}} ${\rm N}_1$ ${\rm Adv}_{cnd}$ ${\rm V}_1$ 过 ${\rm N}_2$ 就好了。\\$\nrightarrow$ ${\rm N}_1$ ${\rm V}_1$ 过 ${\rm N}_2$。 \end{tabular} & \begin{tabular}[c]{@{}l@{}}妹妹如果去过蒙古就好了。\\ If only the younger sister had gone to Mongolia. \\ $\nrightarrow$ 妹妹去过蒙古。\\ The younger sister has gone to Mongolia. \end{tabular} \\\hline
\begin{tabular}[c]{@{}l@{}}Neutral:\\ Choice\end{tabular}  & \begin{tabular}[c]{@{}l@{}}${\rm N}_1$ 和 ${\rm N}_2$， ${\rm PN}$ ${\rm V}_1$ 其中一个。 \\ $\nrightarrow$ ${\rm PN}$ ${\rm V}_1$ ${\rm N}_1$ 。/ ${\rm PN}$ ${\rm V}_1$ ${\rm N}_2$ 。\end{tabular} & \begin{tabular}[c]{@{}l@{}}教授和经理,他喜欢其中一个。\\ He likes either the professor or the manager. \\ $\nrightarrow$ 他喜欢经理。 He likes the manager. \end{tabular} \\\hline
\begin{tabular}[c]{@{}l@{}}Neutral:\\ Argument Drop\end{tabular}  & \begin{tabular}[c]{@{}l@{}} ${\rm N}_1$ 的 ${\rm N}_2$ 在 ${\rm V}_1$ ${\rm N}_3$。 \\ $\nrightarrow$ ${\rm N}_1$ 在 ${\rm V}_1$ ${\rm N}_3$。 \end{tabular} & \begin{tabular}[c]{@{}l@{}}秘书的弟弟在跳舞。 The secretary's younger brother is dancing. \\ $\nrightarrow$ 秘书在跳舞。 The secretary is dancing. \end{tabular} \\\hline
\begin{tabular}[c]{@{}l@{}}Neutral:\\ Drop 要\end{tabular}  & \begin{tabular}[c]{@{}l@{}} 每个 ${\rm N}_1$ 都 要 ${\rm V}_1$ ${\rm N}_2$。 \\ $\nrightarrow$ 每个 ${\rm N}_1$ 都 ${\rm V}_1$ ${\rm N}_2$。 \end{tabular} & \begin{tabular}[c]{@{}l@{}}每个清洁工都要买西瓜。\\ Every cleaner wants to buy watermelon. \\ $\nrightarrow$ 每个清洁工都买西瓜。\\ Every cleaner is going to buy watermelon. \end{tabular} \\\hline
\begin{tabular}[c]{@{}l@{}}Neutral:\\ Adverb Drop\end{tabular}  & \begin{tabular}[c]{@{}l@{}} ${\rm N}_1$ Adv ${\rm V}_1$ 过 ${\rm N}_2$。 \\ $\nrightarrow$ ${\rm N}_1$ ${\rm V}_1$ 过 ${\rm N}_2$。 \end{tabular} & \begin{tabular}[c]{@{}l@{}}清洁工似乎吃过早饭。 The cleaner seems to have eaten breakfast. \\ $\nrightarrow$ 清洁工吃过早饭。The cleaner has eaten breakfast. \end{tabular} \\\hline
\begin{tabular}[c]{@{}l@{}}Neutral:\\ Adverb Drop\end{tabular}  & \begin{tabular}[c]{@{}l@{}} 没法证明 ${\rm N}_1$ ${\rm V}_1$ 过 ${\rm N}_2$。 \\ $\nrightarrow$ ${\rm N}_1$ ${\rm V}_1$ 过 ${\rm N}_2$。 \end{tabular} & \begin{tabular}[c]{@{}l@{}}没法证明爷爷卖过西红柿。\\ It cannot be proven that the grandfather has sold tomatoes. \\ $\nrightarrow$ 爷爷卖过西红柿。\\ The grandfather has sold tomatoes. \end{tabular} \\\bottomrule
\end{tabular}
\end{adjustbox}
\caption{Template Examples of Lexical Overlap Heuristic in Chinese HANS}
\label{tab:app:lexical:overlap}
\end{table*}
\end{CJK*}

\begin{CJK*}{UTF8}{gbsn}
\begin{table*}[t]
\centering
\footnotesize
\scalebox{.9}{
\begin{tabular}{p{2.5cm}p{5cm}p{8cm}}
\toprule
Category & Template & Example \\\midrule
\begin{tabular}[c]{@{}l@{}}Entailment:\\ Adverb Drop\end{tabular}  & \begin{tabular}[c]{@{}l@{}} Adv， ${\rm N}_1$ ${\rm V}_1$ ${\rm N}_2$ 了。 \\ $\rightarrow$ ${\rm N}_1$ ${\rm V}_1$ ${\rm N}_2$ 了。 \end{tabular} & \begin{tabular}[c]{@{}l@{}}反正我们吃橘子了。\\ Anyhow, we ate tangerines. \\ $\rightarrow$ 我们吃橘子了。\\ We ate tangerines. \end{tabular} \\\hline
\begin{tabular}[c]{@{}l@{}}Entailment:\\ Adverb Drop\end{tabular}  & \begin{tabular}[c]{@{}l@{}} Adv ${\rm N}_1$ ${\rm V}_1$ 过 ${\rm N}_2$。 \\ $\rightarrow$ ${\rm N}_1$ ${\rm V}_1$ 过 ${\rm N}_2$。 \end{tabular} & \begin{tabular}[c]{@{}l@{}}果然清洁工听过音乐。\\ As expected, the cleaner has listened to music. \\ $\rightarrow$ 清洁工听过音乐。\\ The cleaner has listened to music. \end{tabular} \\\hline
\begin{tabular}[c]{@{}l@{}}Contradiction:\\ Drop  以为\end{tabular}  & \begin{tabular}[c]{@{}l@{}} ${\rm N}_1$ 以为 ${\rm N}_2$ ${\rm V}_1$ ${\rm N}_3$ 了。 \\ $\nrightarrow$ ${\rm N}_2$ ${\rm V}_1$ ${\rm N}_3$ 了。 \end{tabular} & \begin{tabular}[c]{@{}l@{}}科学家以为法官跳舞了。\\ The scientist thought that the judge danced. \\ $\rightarrow$ 法官跳舞了。\\ The judge danced. \end{tabular} \\\hline
\begin{tabular}[c]{@{}l@{}}Contradiction:\\ Drop\end{tabular}  & \begin{tabular}[c]{@{}l@{}} 谁说 ${\rm N}_1$ 都是 ${\rm V}_1$ ${\rm N}_2$ 的。 \\ $\nrightarrow$ ${\rm N}_1$ 都是 ${\rm V}_1$ ${\rm N}_2$ 的。 \end{tabular} & \begin{tabular}[c]{@{}l@{}}谁说经理都是打领带的。\\ Who told you that managers all wear ties? \\ $\rightarrow$ 经理都是打领带的。\\ Managers all wear ties. \end{tabular} \\\hline
\begin{tabular}[c]{@{}l@{}}Contradiction:\\ Drop\end{tabular}  & \begin{tabular}[c]{@{}l@{}} Num 年后将实现每个 ${\rm N}_1$ 都有${\rm N}_2$。 \\ $\nrightarrow$ 每个 ${\rm N}_1$ 都有${\rm N}_2$。 \end{tabular} & \begin{tabular}[c]{@{}l@{}}三年后将实现每个县都有京剧团。\\ In three years, the goal will be realized that \\ every county has a Chinese operator troupe. \\ $\rightarrow$ 每个县都有京剧团。\\ Every county has a Chinese operator troupe. \end{tabular} \\\hline
\begin{tabular}[c]{@{}l@{}}Neutral: \\ Adv, RC\end{tabular}  & \begin{tabular}[c]{@{}l@{}} Adv ${\rm N}_1$ ${\rm V}_1$ 的 ${\rm N}_2$ ${\rm V}_2$ 过 ${\rm N}_3$。 \\ $\nrightarrow$ ${\rm N}_2$ ${\rm V}_2$ 过 ${\rm N}_3$。 \end{tabular} & \begin{tabular}[c]{@{}l@{}}可能秘书喜欢的艺术家买过哈密瓜。\\ Maybe the artist that the secretary likes \\ has bought Hami melon. \\ $\rightarrow$ 艺术家买过哈密瓜。\\ The artist has bought Hami melon.\end{tabular}  \\\hline
\begin{tabular}[c]{@{}l@{}}Neutral: \\ Drop \end{tabular}  & \begin{tabular}[c]{@{}l@{}} 看是不是 ${\rm N}_1$ ${\rm V}_1$ 的 ${\rm N}_2$。 \\ $\nrightarrow$ 是 ${\rm N}_1$ ${\rm V}_1$ 的 ${\rm N}_2$。 \end{tabular} & \begin{tabular}[c]{@{}l@{}}看是不是领导喜欢的研究生。\\ Let's see if (he/she) is the kind of students the leader likes. \\ $\rightarrow$ 是领导喜欢的研究生。\\ (He/she) is the kind of students the leader likes. \end{tabular} \\\hline
\begin{tabular}[c]{@{}l@{}}Neutral: \\ Drop Adverb \end{tabular}  & \begin{tabular}[c]{@{}l@{}} Adv ${\rm N}_1$ ${\rm V}_1$ ${\rm N}_2$ 了。 \\ $\nrightarrow$ ${\rm N}_1$ ${\rm V}_1$ ${\rm N}_2$ 了。 \end{tabular} & \begin{tabular}[c]{@{}l@{}}也许经理听歌剧了。\\ Maybe the manager listened to the operator. \\ $\rightarrow$ 经理听歌剧了。\\ The manager listened to the operator. \end{tabular} \\\hline
\begin{tabular}[c]{@{}l@{}}Neutral: \\ 连 \end{tabular}  & \begin{tabular}[c]{@{}l@{}} 连 ${\rm V}_1$ ${\rm N}_1$ 的 ${\rm N}_2$ 都 觉得 ADJ。 \\ $\nrightarrow$  ${\rm N}_2$ 都 觉得 ADJ。 \end{tabular} & \begin{tabular}[c]{@{}l@{}}连听昆曲的清洁工都觉得早。\\ Even the cleaners who listen to the Kun opera \\ thinks it's too early. \\ $\rightarrow$ 清洁工都觉得早。\\ Even cleaners think it's too early. \end{tabular} \\\bottomrule
\end{tabular}
}
\caption{Template Examples of Sub-sequence Heuristic in Chinese HANS}
\label{tab:app:subseq:overlap}
\end{table*}
\end{CJK*}

\end{document}